\documentclass[10pt,twocolumn,letterpaper]{article}

\usepackage{wacv}
\usepackage{multirow}
\usepackage{array}
\usepackage{booktabs}
\usepackage{colortbl}
\usepackage{pifont}
\usepackage{adjustbox}

\definecolor{wacvblue}{rgb}{0.21,0.49,0.74}
\usepackage[pagebackref,breaklinks,colorlinks,allcolors=wacvblue]{hyperref}

\def\wacvPaperID{1765}
\def\confName{WACV}
\def\confYear{2027}

\title{GRIT: Gradient Masking for Visually Misgrounded Tokens \\ in Text-Rich Reasoning}

\author{Chanseul Cho \quad Noori Bae \quad Dongheon Lee \quad Youngmin Ro\textsuperscript{*} \\
Machine Intelligence Laboratory, University of Seoul \\
Data Science Interdisciplinary Major\\
\url{https://github.com/chanseul01/GRIT}
}

\begin{document}
\AddToHookNext{shipout/background}{%
  \put(0,-\paperheight){\makebox[\paperwidth]{\raisebox{0.3in}{\parbox{0.8\paperwidth}{\centering\scriptsize
    \copyright~2027 IEEE. Personal use of this material is permitted. Permission from IEEE must be obtained for all other uses, in any current or future media, including reprinting/republishing this material for advertising or promotional purposes, creating new collective works, for resale or redistribution to servers or lists, or reuse of any copyrighted component of this work in other works.}}}}}
\maketitle
\renewcommand{\thefootnote}{\fnsymbol{footnote}}
\footnotetext[1]{Corresponding author.}
\renewcommand{\thefootnote}{\arabic{footnote}}
\begin{abstract}

Reinforcement learning with verifiable rewards (RLVR) has recently been extended to vision-language models (VLMs), yet its effectiveness remains limited for text-rich reasoning tasks. We highlight a key factor contributing to this limitation: models frequently arrive at correct final answers despite misperceiving visual information during intermediate steps. We define this phenomenon as Correct-but-Visually-Misgrounded Reasoning (CVMR). Because standard RLVR rewards only final-answer correctness, it fails to differentiate between properly grounded and misgrounded reasoning, reinforcing both equally. Through token-level analysis, we reveal that CVMR tokens exhibit distinct attention patterns that differ systematically from well-grounded ones. Based on this observation, we propose GRIT, a training framework that incorporates visual grounding into RLVR to mitigate the impact of CVMR. Furthermore, to address the lack of text-rich reasoning benchmarks that allow for systematic evaluation across text length and reasoning depth, we introduce GRIT-Bench. Experiments demonstrate that GRIT significantly improves both reasoning accuracy and grounding fidelity over existing RLVR methods.
\end{abstract}

\section{Introduction}
\label{sec:intro}
Driven by the success of Reinforcement Learning with Verifiable Rewards (RLVR)-based reasoning training in large language models (LLMs)~\citep{deepseek-r1, deepseekmath, kimi-k15, dapo}, there has been growing interest in extending these approaches to vision-language models (VLMs)~\citep{vision-r1, mm-eureka, vl-rethinker, vlaa-thinker, vision-sr1, r1-onevision, skywork-r1v}. These methods have demonstrated significant performance improvements across various visual reasoning tasks. Beyond general visual perception, text-rich reasoning, which involves reading and reasoning over text in images such as financial reports, medical documents, and charts, has emerged as a key capability of VLMs~\citep{textvqa, docvqa, chartqa, docthinker}. These tasks require multi-step reasoning far exceeding simple text recognition, integrating textual content with visual context~\citep{ocr-reasoning, chartqa, charxiv, contextual}.

However, RLVR-driven improvements do not consistently extend to this domain. Despite numerous recent efforts to enhance the visual reasoning capabilities of VLMs through RL~\citep{vision-r1, mm-eureka, vl-rethinker, vlaa-thinker, vision-sr1, r1-onevision, skywork-r1v}, there has been relatively limited systematic analysis of text-rich reasoning. For example, \citet{ocr-reasoning} recently reported that RL-trained models perform even worse than baselines on text-rich reasoning tasks, and we observe similar degradation on text-rich reasoning benchmarks, including OCR-Reasoning, in our experiments.

\begin{figure}[t]
\centering
\includegraphics[width=1.0\linewidth]{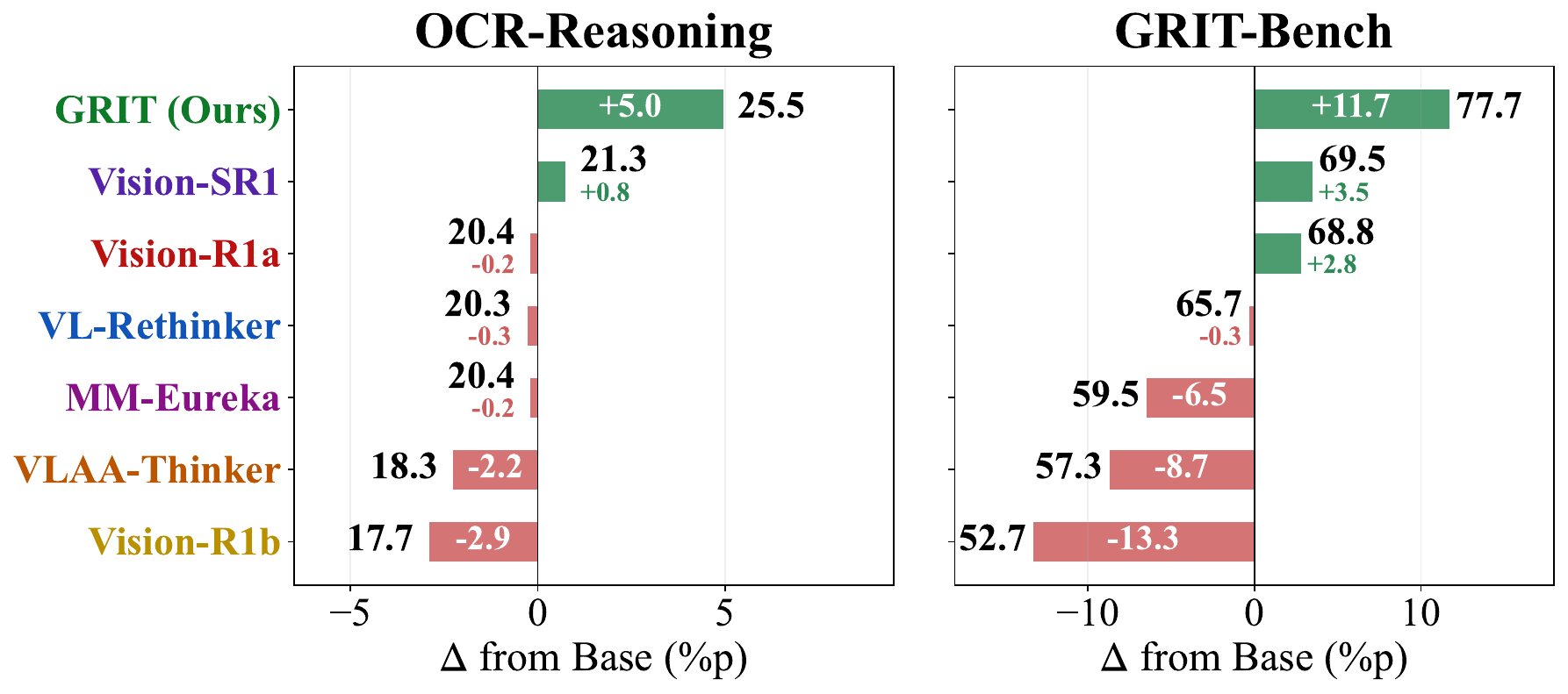}
\caption{Performance change over the base model on text-rich reasoning benchmarks. Existing RLVR methods mostly degrade or show marginal improvement, while GRIT achieves substantial gains.}
\label{fig:teaser}
\vspace{-2mm}
\end{figure}

\begin{figure*}[t]
\centering
\includegraphics[width=1.0\linewidth]{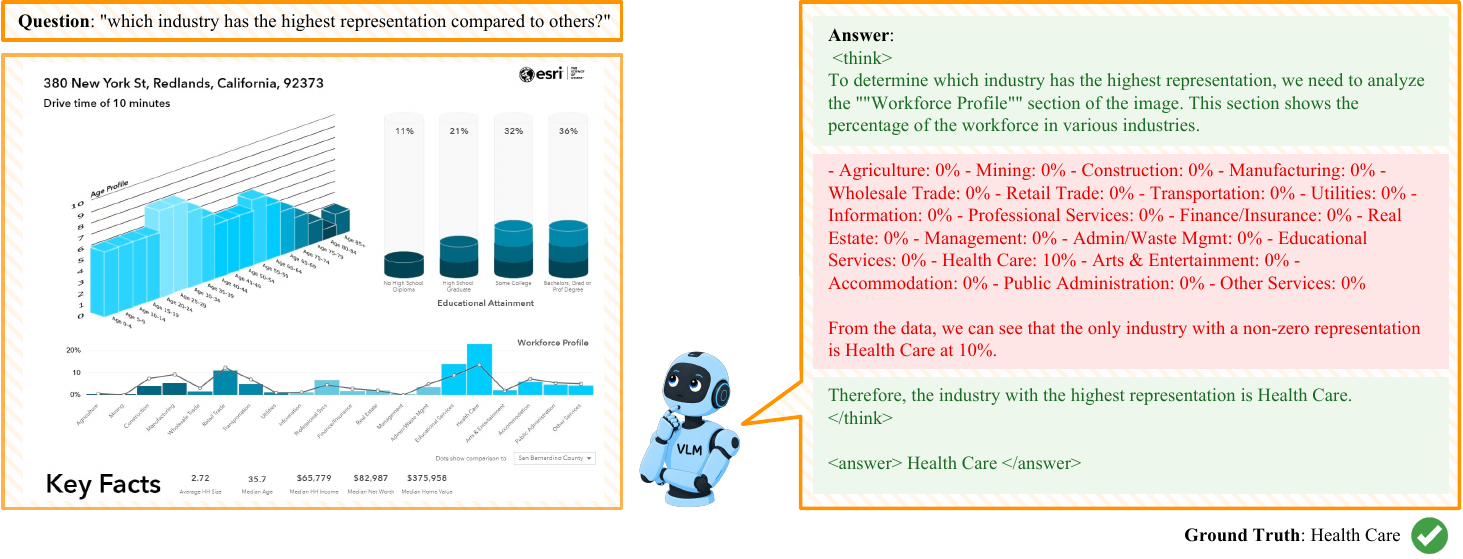}
\caption{An example of Correct-but-Visually-Misgrounded Reasoning (CVMR). The model correctly identifies the relative ranking, noting that Health Care has the highest representation in the infographic, and arrives at the correct answer. However, despite multiple industries
having meaningful percentages, the model misreads all values including Health Care itself. Red-highlighted text
indicates visually misgrounded reasoning.}
\label{fig:cvmr-example}
\vspace{-2mm}
\end{figure*}

To investigate the underlying causes of this degradation, we closely examine VLM responses and identify a critical phenomenon we term Correct-but-Visually-Misgrounded Reasoning (CVMR), where models frequently arrive at correct answers despite misperceiving visual information (additional examples in Supplementary Sec.~\ref{app:qualitative}). For instance, in Figure~\ref{fig:cvmr-example}, the model correctly infers that Health Care has the highest representation to answer the question, yet misreads the specific values of all industries. Unlike standard hallucinations~\citep{hallucination-survey}, this failure mode bypasses answer-only verification because the final output remains correct. Text-rich reasoning tasks are structurally vulnerable to CVMR due to their dense visual elements. Our analysis shows that CVMR occurs at a disproportionately high rate in these tasks (Section~\ref{sec:cvmr}), suggesting that standard RLVR, which primarily relies on answer correctness, creates a systematic risk of reinforcing misgrounded reasoning.

To address this, we propose \textbf{GRIT} (\textbf{Gr}adient Masking for V\textbf{i}sually Misgrounded \textbf{T}okens). We observe that CVMR tokens exhibit weak or diffuse attention over relevant image regions. To capture such attention patterns, GRIT selects a subset of attention heads that sensitively reflect the image regions attended by each generated token, termed \textit{Visual Tracker Heads}. During training, GRIT analyzes the attention patterns from these heads to identify visually misgrounded tokens and masks their gradients to prevent CVMR from being reinforced. To incorporate the faithfulness of visual grounding into the training signal, we introduce an \textit{attention reward} based on their attention mass and entropy. Consequently, as shown in Figure~\ref{fig:teaser}, while existing RLVR methods largely degrade or show marginal improvement on text-rich reasoning benchmarks, GRIT achieves substantial gains.

Furthermore, systematically evaluating the extent to which models rely on correct visual information requires more than answer-level supervision; it requires verifying whether each reasoning step is faithfully grounded in the visual input. Existing text-rich reasoning benchmarks provide valuable testbeds, but they rarely offer the fine-grained reference information needed to assess visual grounding throughout the reasoning process. To enable such analysis, we introduce \textbf{GRIT-Bench}, a benchmark constructed by rendering well-established text-based QA datasets into images. This design yields two advantages. First, it provides a controlled testbed for explicitly measuring the grounding fidelity of VLMs in text-rich reasoning. Second, because each visual input is paired with its exact source text, an LLM judge can evaluate the VLM's reasoning process directly against this text, providing a more reliable basis than image-only evaluation for assessing whether visual information was correctly perceived.

Our contributions are as follows:
\begin{itemize}
    \item We define CVMR (Correct-but-Visually-Misgrounded Reasoning), demonstrate its high prevalence in text-rich reasoning, and show that CVMR tokens exhibit distinct attention patterns.
    \item We propose GRIT, a training framework that mitigates the reinforcement of CVMR through attention-based gradient masking and tailored reward design, achieving substantial improvements over existing RLVR methods on text-rich reasoning tasks.
    \item We construct GRIT-Bench, a text-rich reasoning benchmark comprising approximately 3K samples, which explicitly enables the evaluation of grounding fidelity and fine-grained analysis by text length and reasoning depth.
\end{itemize}

\section{Related Work}
\subsection{RLVR for Vision-Language Models}
Reinforcement learning with verifiable rewards (RLVR) has achieved notable success in enhancing the reasoning capabilities of LLMs~\citep{deepseek-r1, deepseekmath, kimi-k15}. Motivated by this, recent work has extended RLVR to VLMs. Vision-R1a~\citep{vision-r1} applies GRPO-based training with a synthetically constructed multimodal CoT dataset and introduces progressive thinking suppression to reduce overthinking. Vision-R1b~\citep{vision-r1-star} proposes vision-guided reinforcement learning with progressive rule refinement for human-free alignment. MM-Eureka~\citep{mm-eureka} demonstrates that minimalist rule-based rewards combined with a curated multimodal mathematics dataset are sufficient for effective multimodal reasoning. VL-Rethinker~\citep{vl-rethinker} proposes forced rethinking, which appends rethinking triggers to rollouts to encourage explicit self-reflection during RL training. VLAA-Thinker~\citep{vlaa-thinker} investigates the SFT-then-RL paradigm and finds that direct RL with a mixed reward module integrating perception and cognition signals outperforms the conventional pipeline. Vision-SR1~\citep{vision-sr1} proposes a self-rewarding framework that decomposes reasoning into visual perception and language reasoning, using the model itself to evaluate its own visual perception quality. While these methods have demonstrated improvements on general visual reasoning benchmarks, their effectiveness on text-rich reasoning remains limited, as we observe in our experiments. GRIT addresses this gap by incorporating visual grounding into the RLVR training process.

\subsection{Text-Rich Visual Reasoning}
Existing benchmarks for text-rich image understanding, such as TextVQA~\citep{textvqa}, DocVQA~\citep{docvqa}, and ChartQA~\citep{chartqa}, have contributed to evaluating text recognition and comprehension in various visual contexts. However, these benchmarks are largely centered on textual content extraction, where answers can often be directly retrieved from the OCR results of the image. OCR-Reasoning~\citep{ocr-reasoning} addresses this gap by constructing a benchmark where answers are minimally present in the OCR output, requiring the model to perform reasoning rather than extraction. It systematically defines six core reasoning sub-abilities for text-rich scenarios, including spatial, numerical, mathematical, logical, enumerative, and multidisciplinary reasoning, providing a structured evaluation framework that was previously absent. These characteristics make OCR-Reasoning a useful testbed for examining whether RL-based multimodal reasoning methods improve text-rich reasoning beyond mere text extraction. ChartQA-PRO~\citep{chartqapro} and CharXiv~\citep{charxiv} also evaluate chart and figure understanding, which partially involves text-rich reasoning. GRIT-Bench complements these benchmarks by enabling fine-grained analysis across text length and reasoning depth while additionally measuring reasoning fidelity.

\section{CVMR Analysis}
\label{sec:cvmr}
\begin{figure}[t]
\centering
\includegraphics[width=0.8\linewidth]{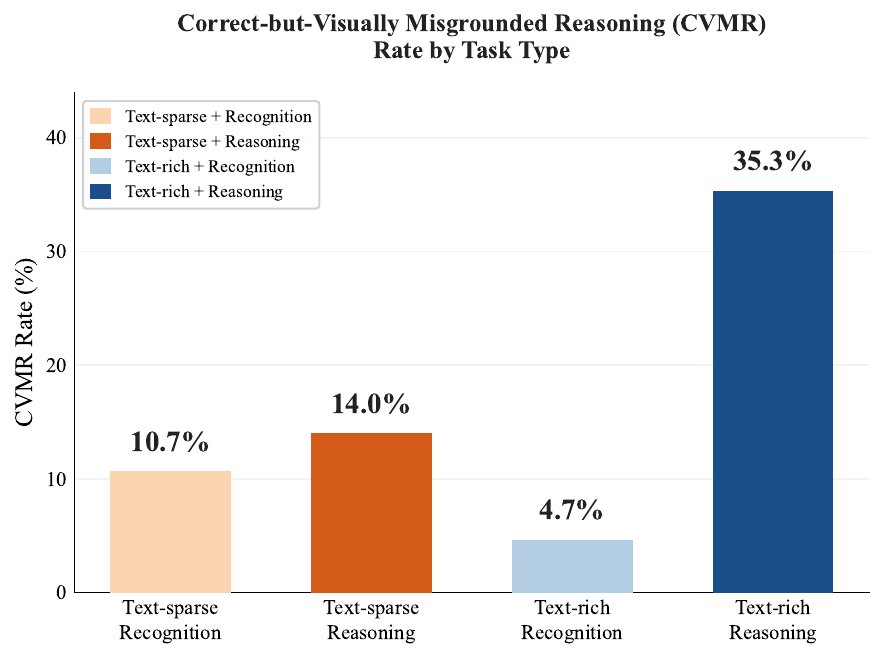}
\caption{CVMR rate among correct responses by task type. Text-rich reasoning exhibits a substantially higher CVMR rate compared to other task types.}
\label{fig:cvmr-rate}
\vspace{-2mm}
\end{figure}
As introduced in Section~\ref{sec:intro}, CVMR refers to the phenomenon where a model arrives at the correct final answer despite misperceiving visual information in the image. In this section, we quantitatively analyze which task types are most susceptible to CVMR and investigate the distinct attention patterns exhibited by tokens associated with CVMR.

To systematically analyze the occurrence of CVMR, we collect 600 samples across four groups defined by two axes: text density and task type, with 150 correct samples per group. Since CVMR by definition occurs exclusively in correct responses, we restrict our samples accordingly. Images containing 200 or more characters of text are classified as text-rich, and those with fewer as text-sparse. Tasks requiring at least two reasoning steps to solve are classified as reasoning, while those answerable by simply locating a value in the image are classified as recognition. Samples are drawn from ChartQA~\citep{chartqa}, ChartQAPro~\citep{chartqapro}, DocVQA~\citep{docvqa}, ScienceQA~\citep{scienceqa}, A-OKVQA~\citep{a-okvqa}, and RealWorldQA~\citep{realworldqa}, with responses generated by Qwen2.5-VL-7B~\cite{qwen25vl}. CVMR is determined for each sample by first using GPT-5~\citep{gpt-5} to judge whether the model's reasoning process contains misperceptions of visual information, followed by human double-verification.

Figure~\ref{fig:cvmr-rate} shows the CVMR rate among correct responses by task type. The CVMR rate in text-rich reasoning is 35.3\%, substantially higher than that of text-sparse reasoning (14.0\%), text-sparse recognition (10.7\%), and text-rich recognition (4.7\%). This result stems from the structural characteristics of text-rich reasoning. Text-rich reasoning tasks contain numerous visual elements that must be read and interpreted, and misperceptions of elements not directly tied to the final answer during multi-step reasoning can still yield correct responses. In contrast, text-rich recognition exhibits the lowest CVMR rate: the retrieved value itself constitutes the answer without multi-step reasoning, meaning any misreading directly leads to an incorrect answer. Because standard RLVR primarily relies on answer correctness as its reward signal, it cannot distinguish visually misgrounded tokens from properly grounded ones. Consequently, tasks with higher CVMR rates risk having misgrounded reasoning reinforced alongside correct ones.

\begin{figure}[t]
\centering
\includegraphics[width=1.0\linewidth]{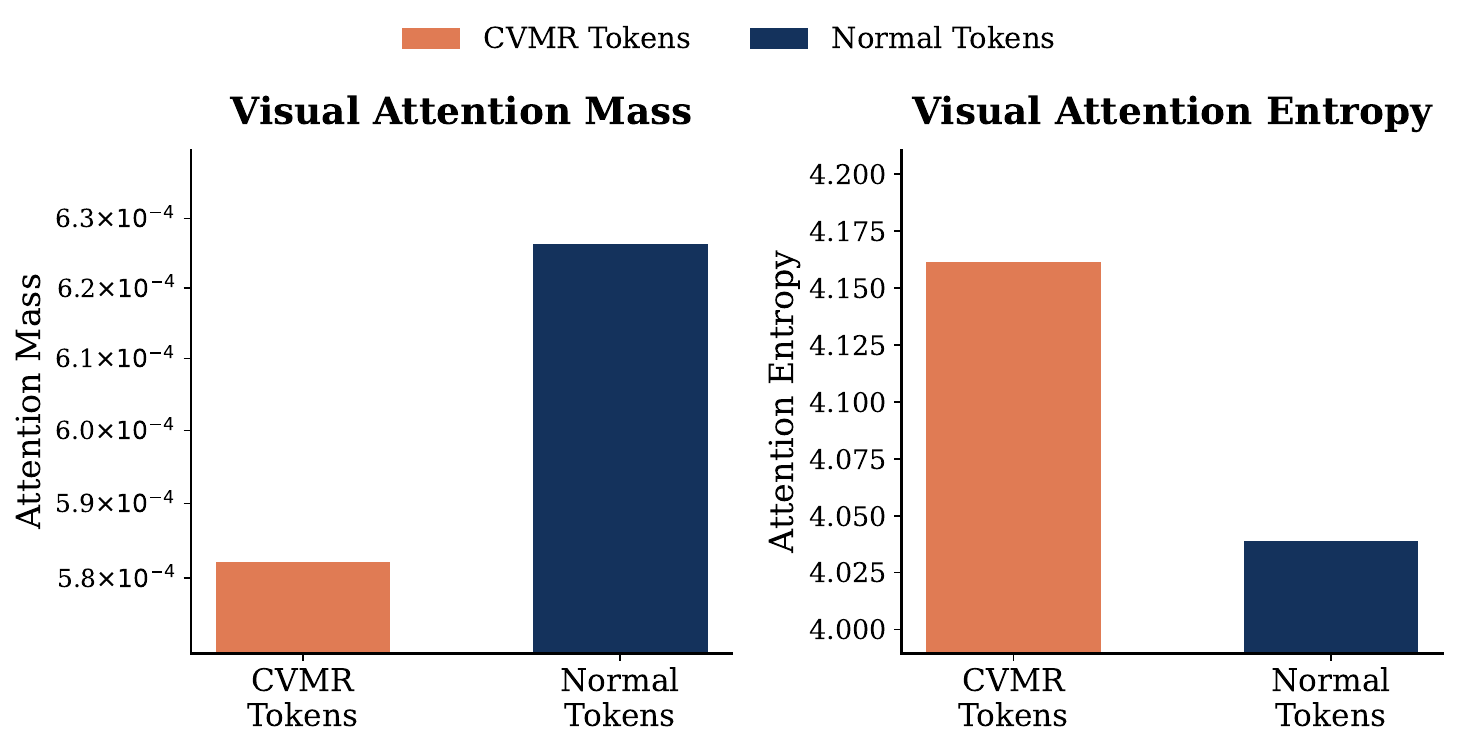}
\caption{Visual attention mass and entropy of CVMR tokens vs.\ normal tokens, measured on Visual Tracker Heads. CVMR tokens show lower attention mass and higher attention entropy (two-sample $z$-test, both $p$-values $< 10^{-4}$).}
\label{fig:attention-cvmr}
\vspace{-2mm}
\end{figure}

We further compare the attention characteristics of text tokens where visual misperceptions occur, which we term CVMR tokens, against normal tokens that correctly reflect visual information. Specifically, we utilize Visual Tracker Heads (Section~\ref{sec:vth}) to analyze two metrics: visual attention mass and visual attention entropy. Visual attention mass represents the total attention a generated text token allocates to image tokens, while visual attention entropy captures whether the attention is focused on specific image regions or spread broadly. For example, a higher attention mass indicates that the model references the image more, while a lower attention entropy indicates more focused attention on specific image regions~\citep{few-attn-heads, why-lvlm-hallucinate-longer}. As shown in Figure~\ref{fig:attention-cvmr}, we find that CVMR tokens exhibit a distinct attention pattern compared to normal tokens, characterized by lower visual attention mass and higher attention entropy (see Supplementary Sec.~\ref{app:qualitative} for qualitative examples). This pattern is consistent across both metrics, suggesting that CVMR tokens do not sufficiently focus on relevant image regions. This observation shows that attention patterns can serve as a meaningful signal for distinguishing CVMR tokens, motivating the design of GRIT.

\section{GRIT}
\label{sec:grit}

In Section~\ref{sec:cvmr}, we observed that CVMR tokens exhibit lower visual attention mass and higher attention entropy. Since RLVR assigns the same reward to all tokens within a correct response, it cannot distinguish visually grounded tokens from misgrounded ones, causing CVMR to be reinforced alongside correct reasoning. In this section, we introduce GRIT, a training framework that addresses this limitation. GRIT consists of three components: (1) Visual Tracker Heads, which select attention heads that reflect token-level variation in visual attention patterns, (2) gradient masking, which blocks gradients for misgrounded tokens based on the attention patterns of selected heads, and (3) attention reward, which incorporates the faithfulness of visual grounding into the reward signal.

\subsection{Group Relative Policy Optimization}
\label{sec:grpo}

We adopt GRPO~\citep{deepseekmath} as the base RL algorithm. GRPO samples $G$ responses for each prompt and converts the relative differences in rewards within the group into advantages for training. Specifically, given an image-question pair $(I, q)$ and $G$ responses $\{o_i\}_{i=1}^{G}$ with rewards $\{R_i\}_{i=1}^{G}$, the advantage assigned to each token $t$ of response $o_i$ is:
\begin{equation}
    \hat{A}_{i,t} = \frac{R_i - \mathrm{mean}(\{R_i\}_{i=1}^{G})}{\mathrm{std}(\{R_i\}_{i=1}^{G})}
\end{equation}
where the same normalized advantage is shared across all tokens within a response. The policy is updated to maximize:
{\small
\begin{multline}
    \mathcal{J}_{\mathrm{GRPO}}(\theta) = \frac{1}{G} \sum_{i=1}^{G} \frac{1}{|o_i|} \sum_{t=1}^{|o_i|} \bigg( \min \Big( r_{i,t}(\theta) \hat{A}_{i,t}, \\
    \mathrm{clip}\big(r_{i,t}(\theta), 1\!-\!\epsilon, 1\!+\!\epsilon\big) \hat{A}_{i,t} \Big) - \beta D_{\mathrm{KL}}(\pi_\theta \| \pi_{\mathrm{ref}}) \bigg)
\end{multline}
}%
where the expectation is taken over $ (I,q,a) \sim \mathcal{D} $ and $ \{o_i \} _ {i=1 } ^ {G } \sim \pi _ {\theta _ {\mathrm {old } } } (\cdot \mid I,q) $ , and
\begin{equation}
r_{i,t}(\theta) = \frac{\pi _ \theta (o_{i,t} \mid I, q, o_{i,<t}) }{\pi _ {\theta _ {\mathrm {old } } } (o_{i,t} \mid I, q, o_{i,<t}) }
\end{equation}
is the per-token policy ratio. To allow sufficient exploration during text-rich reasoning training, we remove the KL penalty term, apply asymmetric clipping ($\epsilon_l = 0.2$, $\epsilon_h = 0.28$), and adopt token-level loss normalization~\citep{dapo}.

\subsection{Visual Tracker Heads}
\label{sec:vth}

Recent studies have shown that individual attention heads exhibit specialized roles~\citep{voita2019heads, attn-heads-survey, ocr-heads, few-attn-heads}, suggesting that a targeted subset of heads may be more informative than using all heads uniformly. Since GRIT masks gradients at the token level (Section~\ref{sec:gradmask}), it requires a way to assess whether each individual token is visually grounded. This in turn demands attention heads that sensitively reflect which image region each generated token references. If a head attends to the same image region regardless of which token is being generated, it cannot distinguish well-grounded tokens from misgrounded ones. We select heads satisfying this property and refer to them as \textit{Visual Tracker Heads}. We propose a two-stage procedure for their identification.

\paragraph{Mass Filtering.}
The selection is performed on 1,000 samples drawn from~\citep{chartqa, chartqapro, docvqa, scienceqa, a-okvqa, realworldqa}. Since only a small fraction of attention heads contribute meaningfully to text-image interaction, we first compute the mean visual attention mass across all response tokens for each head, and retain the top 50 heads that effectively attend to image tokens.

\paragraph{Diversity Ranking.}
Among the retained heads, we identify those whose attention distribution over image tokens changes most across generated tokens. Such heads are informative because their token-level variation enables per-token grounding assessment. For each head, we compute the mean attention map $\bar{\mathbf{a}}_h$ across all response tokens, then measure \textit{token-conditioned spatial diversity} as:
\begin{equation}
    \label{eq:diversity}
    \mathrm{div}(h) = \frac{1}{T} \sum_{t=1}^{T} \left( 1 - \mathrm{cos}(\mathbf{a}_{h,t}, \, \bar{\mathbf{a}}_h) \right)
\end{equation}
where $\mathbf{a}_{h,t}$ is the attention map of head $h$ at token $t$, $\mathrm{cos}(\cdot, \cdot)$ denotes cosine similarity, and $T$ is the total number of response tokens in the selection set. A higher diversity score indicates that the head shifts its focus to different image regions depending on which token is being generated, making it suitable for detecting token-level grounding differences. We select the top 5 heads by diversity as Visual Tracker Heads. Further details on Visual Tracker Heads are provided in Supplementary Sec.~\ref{app:vth}.

\subsection{Gradient Masking}
\label{sec:gradmask}

In standard RLVR, all tokens within a response share the same response-level advantage, so visually misgrounded tokens in a positively reinforced correct response receive the same policy credit as grounded tokens. To prevent this, GRIT selectively blocks gradients for tokens that are likely to be visually misgrounded, using the attention patterns of Visual Tracker Heads.

\paragraph{Two-Stage Token Filtering.}
For each token in a correct response, we compute its visual attention mass and entropy on the Visual Tracker Heads. Within each response, tokens suspected of being misgrounded are identified through two successive filters:
\begin{enumerate}
    \item \textbf{Low Mass Filter}: tokens in the bottom 50\% of visual attention mass within the response are selected.
    \item \textbf{High Entropy Filter}: among those, tokens in the top 50\% of visual attention entropy are selected.
\end{enumerate}
The gradients of the resulting tokens are masked. This filtering is intentionally approximate and conservative: masking removes a token's policy-loss contribution at the current update rather than applying a gradient in the opposite direction. This filtering uses attention statistics as a proxy for token-level grounding. Robustness to alternative masking strategies and threshold choices is evaluated in Section~\ref{sec:ablation} and discussed further in Supplementary Sec.~\ref{app:discussion}.

\paragraph{Protected Tokens.}
Not all tokens should be subject to masking. Answer format tags (\texttt{<answer>}, \texttt{<think>}), mathematical operators (\texttt{+\,-\,=\,*\,/}), and punctuation (\texttt{.\,,\,(\,)\,"\,'}) serve structural roles unrelated to visual perception. These \textit{protected tokens} always retain their gradients regardless of their attention patterns.

\paragraph{Selective Application to Correct Responses.}
Gradient masking is applied only to correct responses because CVMR is defined by a correct final answer paired with visually misgrounded reasoning. In these responses, outcome correctness
does not resolve token-level grounding quality. Incorrect responses retain the standard GRPO update without additional masking.

The resulting training objective incorporates a per-token mask $m_{i,t} \in \{0, 1\}$:
{\small
\begin{multline}
    \mathcal{J}_{\mathrm{GRIT}}(\theta) = \frac{1}{\sum_{i=1}^{G} |o_i|} \sum_{i=1}^{G} \sum_{t=1}^{|o_i|} m_{i,t} \cdot \min \Big( r_{i,t}(\theta) \hat{A}_{i,t}, \\
    \mathrm{clip}\big(r_{i,t}(\theta), 1\!-\!\epsilon_l, 1\!+\!\epsilon_h\big) \hat{A}_{i,t} \Big)
\end{multline}
}%
where $m_{i,t} = 0$ for tokens masked by Two-Stage Token Filtering in correct responses, and $m_{i,t} = 1$ otherwise.

\subsection{Attention Reward}
\label{sec:attnreward}

Gradient masking suppresses the reinforcement of misgrounded tokens at the token level, but does not reflect differences in visual grounding quality across responses. To complement this, we additionally introduce an attention reward that incorporates response-level grounding quality into the reward signal, using the attention patterns of Visual Tracker Heads.

\paragraph{Attention Score.}
For each response, we compute the mean visual attention mass and mean visual attention entropy across all response tokens based on Visual Tracker Heads. As observed in Section~\ref{sec:cvmr}, visually well-grounded tokens exhibit higher attention mass and lower entropy, so we define the per-response attention score as:
\begin{equation}
    s_i = \log(\overline{M}_i) - \log(\overline{E}_i)
\end{equation}
where $\overline{M}_i$ and $\overline{E}_i$ are the mean visual attention mass and entropy of response $o_i$, respectively. This score is normalized within the group and mapped to $(0, 1)$ via sigmoid:
{\small
\begin{gather}
    z_i = \frac{s_i - \mathrm{mean}(\{s_j\}_{j=1}^{G})}{\mathrm{std}(\{s_j\}_{j=1}^{G})} \\[4pt]
    R^{\mathrm{attn}}_i = \sigma(z_i) \cdot c_i
\end{gather}
}%
where $c_i \in \{0, 1\}$ is the correctness indicator of response $o_i$. We gate the attention reward by correctness so that a high attention score cannot compensate for an incorrect final answer. The reward therefore differentiates grounding quality only among correct responses.

\paragraph{Final Reward.}
The final reward combines the correctness reward, format reward, and attention reward, and serves as the reward used for GRPO advantage computation:
\begin{equation}
    \tilde{R}_i = R^{\mathrm{format}}_i + R^{\mathrm{correct}}_i + R^{\mathrm{attn}}_i
\end{equation}
where $R^{\mathrm{format}}_i$ is a binary reward for whether the response follows the designated output format (\texttt{<think>}, \texttt{<answer>} tags) and $R^{\mathrm{correct}}_i$ is a binary reward based on answer correctness.

\section{GRIT-Bench}
\label{sec:grit-bench}

Existing text-rich benchmarks are primarily designed around OCR and perception, and only a few such as OCR-Reasoning~\citep{ocr-reasoning} aim to systematically evaluate text-rich reasoning. In particular, no benchmark allows fine-grained analysis of model performance by text length or reasoning depth.

This is because constructing problems that are text-rich and require multi-step reasoning is inherently difficult, and ground truth verification is challenging~\citep{ocr-reasoning}. GRIT-Bench addresses this by rendering text-based QA datasets into images. The original source text is preserved prior to rendering. During evaluation, the judge (GPT-5 mini) receives this source text rather than the image to analyze the model's reasoning process. This enables direct measurement of whether the model correctly read the information in the image.

Specifically, well-established table and document QA datasets~\citep{finqa, tatqa, hitab, drop, wtq, convfinqa, multihiertt} are rendered into images for use as VLM inputs. Text lengths range from 238 to 42,343 characters (mean 3,136), and samples with excessive text are split into multiple images to avoid resolution issues. For Reasoning Fidelity (RF) evaluation, key reasoning steps for each sample are extracted by GPT-5 from the source text. Human double-checking is performed to verify both that the answer is derivable from the rendered image and that the reasoning steps are correct.

GRIT-Bench evaluates with two metrics: Accuracy and Reasoning Fidelity (RF). Accuracy measures final answer correctness. RF evaluates the visual grounding faithfulness of the reasoning process, defined as:
\begin{equation}
    \mathrm{RF} = \frac{n_{\mathrm{faithful}}}{n_{\mathrm{total}} + n_{\mathrm{ungrounded}}}
\end{equation}
where $n_{\mathrm{faithful}}$ is the number of key reasoning steps that the model attempted and that are consistent with the source text, $n_{\mathrm{total}}$ is the total number of key reasoning steps, and $n_{\mathrm{ungrounded}}$ is the number of claims not supported by the source text beyond the key steps. A higher RF indicates that the reasoning process more accurately reflects the information in the image and less frequently asserts content absent from the source.

GRIT-Bench classifies difficulty along two axes: text length and reasoning depth. Text length is divided into Short ($<$1K), Medium (1--2K), and Long (2K+), and reasoning depth into Simple (2 steps), Moderate (3), and Complex (4+). The two axes form 9 difficulty combinations over 2,855 samples (Table~\ref{tab:grit-bench-dist}). Details on dataset composition are provided in Supplementary Sec.~\ref{app:grit-bench}.

GRIT-Bench provides a controlled diagnostic setting for reasoning fidelity and difficulty analysis. We complement this evaluation with external text-rich benchmarks, including OCR-Reasoning~\cite{ocr-reasoning}, ChartQA-PRO~\cite{chartqapro}, and CharXiv~\cite{charxiv}. Further discussion of the benchmark scope is provided in Supplementary Sec.~\ref{app:discussion}.

\begin{table}[t]
\centering
\caption{Distribution of GRIT-Bench samples across difficulty combinations.}
\label{tab:grit-bench-dist}
\resizebox{0.8\columnwidth}{!}{%
\begin{tabular}{l|ccc|c}
\toprule
& \textbf{Simple} & \textbf{Moderate} & \textbf{Complex} & \textbf{Total} \\
\midrule
Short ($<$1K) & 198 & 396 & 169 & 763 \\
Medium (1--2K) & 210 & 476 & 270 & 956 \\
Long (2K+) & 263 & 540 & 333 & 1,136 \\
\midrule
Total & 671 & 1,412 & 772 & 2,855 \\
\bottomrule
\end{tabular}
}
\end{table}

\section{Experiments}
\label{sec:experiments}
\begin{table*}[t]
\centering
\caption{Results across text-rich reasoning and general benchmarks. Best results are in \textbf{bold}, second best are \underline{underlined}. $\Delta$ denotes the difference from the base model.}
\label{tab:combined-results}
\resizebox{\textwidth}{!}{%
\begin{tabular}{l|cccccc|cc|c|cc|c|c}
\toprule
& \multicolumn{8}{c|}{\textbf{Text-Rich Reasoning}} & \multicolumn{3}{c|}{\textbf{Chart Understanding}} & \multicolumn{2}{c}{\textbf{General}} \\
\cmidrule(lr){2-9} \cmidrule(lr){10-12} \cmidrule(lr){13-14}
& \multicolumn{8}{c|}{\textbf{OCR-Reasoning}} & \multirow{2}{*}{\textbf{ChartQA-P}} & \multicolumn{2}{c|}{\textbf{CharXiv}} & \multirow{2}{*}{\textbf{MMStar}} & \multirow{2}{*}{\textbf{HalluB.}} \\
\cmidrule(lr){2-9} \cmidrule(lr){11-12}
\multirow{-3}{*}{\textbf{Model}}
& \textbf{Spatial} & \textbf{Numerical} & \textbf{Math} & \textbf{Enum.} & \textbf{Logical} & \textbf{Multi.} & \textbf{Overall} & $\boldsymbol{\Delta}$
& & \textbf{Reas.} & \textbf{Desc.} & & \\
\midrule
\multicolumn{14}{c}{\textit{Based on Qwen2.5-VL-3B}} \\
\midrule
Qwen2.5-VL-3B & 9.17 & 10.55 & 10.78 & 20.11 & 8.33 & 13.14 & 12.07 & -- & 29.8 & 29.7 & \underline{65.3} & 52.9 & \underline{38.5} \\
GRPO-3B & 15.60 & 9.05 & 8.82 & 18.44 & 9.72 & 13.87 & 11.97 & \textcolor{red}{-0.10} & \underline{30.0} & \underline{29.8} & \underline{65.3} & \underline{53.2} & 38.2 \\
DAPO-3B & 11.01 & 8.79 & 9.80 & 20.11 & 9.72 & 12.41 & 11.60 & \textcolor{red}{-0.47} & 30.3 & 29.4 & 65.2 & 53.1 & 37.8 \\
\rowcolor{gray!20}
\textbf{GRIT-3B (Ours)} & \textbf{15.60} & \textbf{12.06} & \textbf{12.75} & \textbf{23.46} & \textbf{15.97} & \textbf{16.06} & \textbf{15.43} & \textcolor{green!50!black}{\textbf{+3.36}} & \textbf{30.6} & \textbf{30.0} & \textbf{65.5} & \textbf{53.6} & \textbf{39.3} \\
\midrule
\multicolumn{14}{c}{\textit{Based on Qwen2.5-VL-7B}} \\
\midrule
Qwen2.5-VL-7B & 11.93 & 22.11 & 14.71 & 27.93 & 14.58 & 24.09 & 20.58 & -- & 38.5 & 38.1 & 71.2 & 61.5 & 47.3 \\
GRPO-7B & 12.84 & 20.60 & 11.76 & 26.82 & 11.11 & 22.63 & 18.99 & \textcolor{red}{-1.59} & 38.1 & 37.6 & 71.4 & 61.5 & \underline{47.5} \\
DAPO-7B & 12.84 & 19.35 & 12.75 & 27.37 & 11.81 & \textbf{28.47} & 19.55 & \textcolor{red}{-1.03} & 38.3 & 37.4 & 71.2 & 61.8 & 46.4 \\
Vision-R1b-7B & 14.68 & 16.58 & 14.71 & 26.26 & 9.03 & 23.36 & 17.68 & \textcolor{red}{-2.90} & 37.5 & 36.5 & 71.0 & 61.2 & 47.0 \\
VLAA-Thinker-7B & 11.01 & 20.10 & 12.75 & 27.37 & 9.72 & 20.44 & 18.33 & \textcolor{red}{-2.25} & 32.3 & 37.3 & 70.6 & 62.3 & 45.9 \\
MM-Eureka-7B & 12.84 & 20.35 & 15.69 & 27.93 & 12.50 & \textbf{28.47} & 20.39 & \textcolor{red}{-0.19} & 36.7 & \underline{39.3} & 71.3 & \underline{62.6} & 46.4 \\
VL-Rethinker-7B & \underline{15.60} & 19.35 & \textbf{19.61} & 25.70 & 15.97 & 24.82 & 20.30 & \textcolor{red}{-0.28} & \underline{40.4} & 37.7 & 69.9 & \underline{62.6} & 46.2 \\
Vision-R1a-7B & 10.09 & 19.60 & 12.75 & 29.61 & \underline{19.44} & 25.55 & 20.39 & \textcolor{red}{-0.19} & 39.3 & 37.1 & 64.5 & 58.6 & 42.6 \\
Vision-SR1-7B & 13.76 & \underline{22.36} & 13.73 & \underline{31.28} & 13.89 & 24.82 & \underline{21.33} & \textcolor{green!50!black}{+0.75} & 39.8 & 39.0 & \textbf{72.2} & 62.5 & 47.3 \\
\rowcolor{gray!20}
\textbf{GRIT-7B (Ours)} & \textbf{18.35} & \textbf{26.88} & \underline{16.67} & \textbf{32.96} & \textbf{22.92} & \underline{27.01} & \textbf{25.54} & \textcolor{green!50!black}{\textbf{+4.96}} & \textbf{41.1} & \textbf{39.4} & \underline{72.0} & \textbf{63.4} & \textbf{47.7} \\
\bottomrule
\end{tabular}
}
\end{table*}

\begin{table*}[t]
\centering
\caption{GRIT-Bench results across text length and reasoning depth. Acc: Accuracy (\%), RF: Reasoning Fidelity (\%) = faithful / (total\_steps + ungrounded\_claims). Best results are in \textbf{bold}, second best are \underline{underlined}. $\Delta$ denotes the difference from the base model.}
\label{tab:grit-bench}
\resizebox{\textwidth}{!}{%
\begin{tabular}{l|cc cc cc|cc cc cc|cc cc cc|cc|cc}
\toprule
& \multicolumn{6}{c|}{\textbf{Short ($<$1K)}} & \multicolumn{6}{c|}{\textbf{Medium (1--2K)}} & \multicolumn{6}{c|}{\textbf{Long (2K+)}} & \multicolumn{2}{c|}{\multirow{2}{*}{\textbf{Overall}}} & \multicolumn{2}{c}{\multirow{2}{*}{$\boldsymbol{\Delta}$}} \\
\cmidrule(lr){2-7} \cmidrule(lr){8-13} \cmidrule(lr){14-19}
& \multicolumn{2}{c}{Simple} & \multicolumn{2}{c}{Moderate} & \multicolumn{2}{c|}{Complex}
& \multicolumn{2}{c}{Simple} & \multicolumn{2}{c}{Moderate} & \multicolumn{2}{c|}{Complex}
& \multicolumn{2}{c}{Simple} & \multicolumn{2}{c}{Moderate} & \multicolumn{2}{c|}{Complex}
& \multicolumn{2}{c|}{} & \multicolumn{2}{c}{} \\
\cmidrule(lr){2-3} \cmidrule(lr){4-5} \cmidrule(lr){6-7} \cmidrule(lr){8-9} \cmidrule(lr){10-11} \cmidrule(lr){12-13} \cmidrule(lr){14-15} \cmidrule(lr){16-17} \cmidrule(lr){18-19} \cmidrule(lr){20-21} \cmidrule(lr){22-23}
\multirow{-3}{*}{\textbf{Model}}
& Acc & RF & Acc & RF & Acc & RF
& Acc & RF & Acc & RF & Acc & RF
& Acc & RF & Acc & RF & Acc & RF
& Acc & RF & Acc & RF \\
\midrule
\multicolumn{23}{c}{\textit{Based on Qwen2.5-VL-3B}} \\
\midrule
Qwen2.5-VL-3B
& 45.5 & 61.9 & 52.0 & 67.4 & \underline{49.7} & 69.0
& \underline{50.0} & \underline{57.7} & \underline{44.5} & \underline{62.5} & \underline{40.0} & 65.3
& 39.5 & 52.0 & 40.9 & \underline{61.1} & \underline{39.0} & \underline{64.0}
& \underline{44.1} & \underline{62.4} & -- & -- \\
GRPO-3B
& 47.0 & 62.7 & 51.3 & \underline{68.0} & 46.2 & 68.7
& 49.5 & 57.0 & 43.3 & 61.0 & 38.5 & \underline{65.5}
& 37.3 & 49.8 & 38.3 & 57.2 & 36.6 & 62.7
& 42.6 & 61.2 & \textcolor{red}{-1.5} & \textcolor{red}{-1.2} \\
DAPO-3B
& \underline{51.5} & \underline{62.8} & \underline{53.3} & 67.3 & 45.6 & \underline{69.5}
& 44.8 & 56.6 & 44.1 & 62.2 & 37.4 & 65.4
& \underline{39.9} & \underline{52.2} & \underline{40.9} & 58.6 & 36.0 & 62.4
& 43.5 & 61.8 & \textcolor{red}{-0.6} & \textcolor{red}{-0.6} \\
\rowcolor{gray!20}
\textbf{GRIT-3B (Ours)}
& \textbf{55.6} & \textbf{65.1} & \textbf{56.6} & \textbf{70.0} & \textbf{47.9} & \textbf{69.8}
& \textbf{50.0} & \textbf{58.8} & \textbf{49.8} & \textbf{64.5} & \textbf{40.0} & \textbf{66.0}
& \textbf{45.6} & \textbf{55.9} & \textbf{46.5} & \textbf{58.3} & \textbf{40.5} & \textbf{65.0}
& \textbf{48.0} & \textbf{63.4} & \textcolor{green!50!black}{\textbf{+3.9}} & \textcolor{green!50!black}{\textbf{+1.0}} \\
\midrule
\multicolumn{23}{c}{\textit{Based on Qwen2.5-VL-7B}} \\
\midrule
Qwen2.5-VL-7B
& 77.3 & 82.8 & 68.9 & 79.7 & 74.6 & 81.7
& 72.4 & 74.9 & 68.7 & 80.0 & 52.2 & 77.5
& 62.4 & 70.2 & 65.2 & 75.2 & 58.6 & 77.7
& 66.0 & 77.6 & -- & -- \\
GRPO-7B
& 78.8 & 83.4 & 73.0 & 81.5 & 73.4 & 81.6
& 71.4 & 74.6 & 67.0 & 76.8 & 56.3 & 77.9
& 54.8 & 62.3 & 57.6 & 68.4 & 54.1 & 73.5
& 63.9 & 74.8 & \textcolor{red}{-2.1} & \textcolor{red}{-2.8} \\
DAPO-7B
& 78.8 & 82.8 & 74.0 & 81.7 & 72.8 & 82.5
& 71.4 & 75.1 & 68.7 & 78.3 & 55.9 & 78.9
& 54.0 & 62.9 & 57.8 & 68.2 & 57.7 & 73.8
& 64.7 & 75.3 & \textcolor{red}{-1.3} & \textcolor{red}{-2.3} \\
Vision-R1b-7B
& 77.8 & 79.7 & 69.7 & 78.1 & 68.0 & 82.8
& 66.2 & 70.8 & 55.3 & 68.5 & 51.9 & 74.6
& 33.8 & 41.0 & 37.4 & 48.2 & 37.8 & 55.1
& 52.7 & 64.3 & \textcolor{red}{-13.3} & \textcolor{red}{-13.3} \\
VLAA-Thinker-7B
& 77.8 & 85.4 & 71.2 & 84.6 & 69.2 & 82.4
& 65.7 & 71.2 & 60.3 & 76.3 & 55.9 & 77.8
& 40.3 & 48.1 & 46.7 & 59.1 & 44.7 & 65.2
& 57.3 & 71.1 & \textcolor{red}{-8.7} & \textcolor{red}{-6.5} \\
MM-Eureka-7B
& 81.8 & 81.7 & 77.8 & 85.0 & 66.3 & 82.3
& 70.0 & 72.9 & 65.1 & 76.3 & 52.2 & 75.5
& 42.2 & 49.3 & 50.4 & 60.6 & 40.8 & 65.0
& 59.5 & 71.1 & \textcolor{red}{-6.5} & \textcolor{red}{-6.5} \\
VL-Rethinker-7B
& 78.8 & 85.2 & 77.0 & 86.2 & 69.2 & \underline{86.0}
& 70.0 & 78.2 & 67.4 & 79.9 & 55.9 & 78.5
& 59.7 & 66.3 & 62.6 & 73.1 & 55.6 & 73.4
& 65.7 & 77.9 & \textcolor{red}{-0.3} & \textcolor{green!50!black}{+0.3} \\
Vision-R1a-7B
& \underline{82.8} & \underline{87.1} & \underline{82.8} & 86.1 & \underline{71.6} & 81.8
& 71.9 & 74.2 & \underline{73.9} & 79.7 & 60.0 & 72.9
& 65.0 & 69.4 & 61.1 & 69.5 & 55.6 & 71.1
& 68.8 & 76.3 & \textcolor{green!50!black}{+2.8} & \textcolor{red}{-1.3} \\
Vision-SR1-7B
& 79.3 & 82.0 & 81.1 & \underline{87.7} & 68.6 & 82.0
& \underline{73.8} & \underline{79.1} & 73.1 & \underline{82.7} & \underline{62.2} & \underline{82.2}
& \underline{65.4} & \underline{76.0} & \underline{65.6} & \underline{78.8} & 58.0 & 76.4
& \underline{69.5} & \underline{80.9} & \textcolor{green!50!black}{+3.5} & \textcolor{green!50!black}{+3.3} \\
\rowcolor{gray!20}
\textbf{GRIT-7B (Ours)}
& \textbf{89.4} & \textbf{88.1} & \textbf{87.9} & \textbf{87.6} & \textbf{84.0} & \textbf{87.7}
& \textbf{88.6} & \textbf{83.9} & \textbf{81.7} & \textbf{84.0} & \textbf{65.2} & \textbf{82.5}
& \textbf{71.1} & \textbf{78.6} & \textbf{72.8} & \textbf{80.9} & \textbf{66.1} & \textbf{81.7}
& \textbf{77.7} & \textbf{83.5} & \textcolor{green!50!black}{\textbf{+11.7}} & \textcolor{green!50!black}{\textbf{+5.9}} \\
\bottomrule
\end{tabular}
}
\end{table*}

\subsection{Experimental Setup}

For a fair comparison with prior RLVR methods built on the Qwen2.5-VL family, we use Qwen2.5-VL-3B-Instruct and Qwen2.5-VL-7B-Instruct~\cite{qwen25vl}. To verify that GRIT is not tied to a particular backbone, we also apply it to Qwen3-VL-4B-Instruct~\cite{qwen3vl}, a newer generation with a redesigned multimodal architecture (Supplementary Sec.~\ref{sec:qwen3}).

We train on the ViRL 39K dataset used by VL-Rethinker~\citep{vl-rethinker}. We apply GRPO with the DAPO modifications described in Section~\ref{sec:grpo} (KL penalty removal, asymmetric clipping, and token-level loss normalization) with learning rate $1 \times 10^{-6}$. Full training details are provided in Supplementary Sec.~\ref{app:implementation}.

We compare against existing RLVR methods based on Qwen2.5-VL-7B: Vision-SR1~\citep{vision-sr1}, Vision-R1a~\citep{vision-r1}, Vision-R1b~\citep{vision-r1-star}, VL-Rethinker~\citep{vl-rethinker}, MM-Eureka~\citep{mm-eureka}, and VLAA-Thinker~\citep{vlaa-thinker}, evaluated using their officially released weights. We also train GRPO and DAPO with the same dataset and training configuration as GRIT to precisely measure the effect of our method. We evaluate on OCR-Reasoning~\citep{ocr-reasoning}, ChartQA-PRO~\citep{chartqapro}, CharXiv~\citep{charxiv}, MMStar~\citep{mmstar}, HallusionBench~\citep{hallusionbench}, and GRIT-Bench (Section~\ref{sec:grit-bench}).

\subsection{Main Results}
\paragraph{Results on Text-Rich Reasoning and General Benchmarks.} Table~\ref{tab:combined-results} presents results on text-rich reasoning and general benchmarks. On OCR-Reasoning, existing RLVR models mostly degrade or improve only marginally over the base model, while GRIT-7B achieves a substantial $+4.96$ gain. On ChartQA-PRO and CharXiv, which partially contain text-rich reasoning samples, GRIT also records competitive performance. On MMStar, a broad evaluation of general visual capabilities, GRIT achieves the best score among all models. The benefit is therefore not limited to text-rich reasoning. Meanwhile, on HallusionBench (qAcc), which evaluates whether a model responds sensitively to changes in visual conditions for the same question, most existing RLVR methods degrade or fail to improve upon the base model. In contrast, GRIT shows a slight improvement in qAcc, indicating that GRIT is more robust against visual misgrounding compared to standard RLVR methods.

\begin{figure}[t]
\centering
\includegraphics[width=1.0\linewidth]{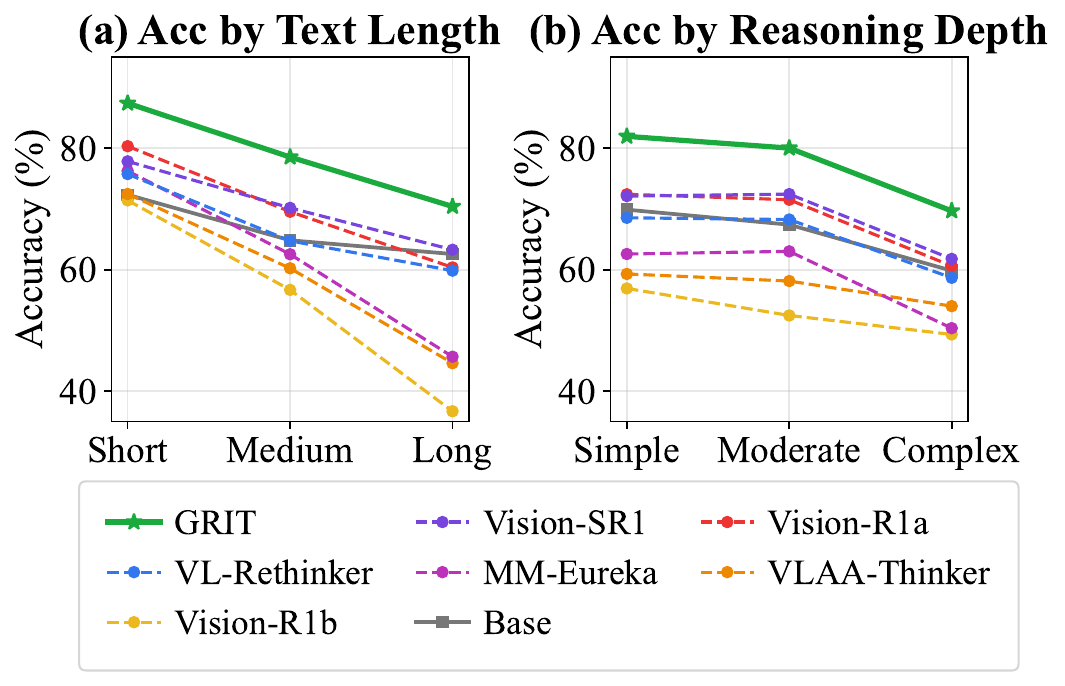}
\caption{GRIT-Bench analysis on 7B models across text length and reasoning depth. (a) Accuracy by text length. (b) Accuracy by reasoning depth.}
\label{fig:grit-bench-analysis}
\end{figure}

\paragraph{Results on GRIT-Bench.}
Table~\ref{tab:grit-bench} and Figure~\ref{fig:grit-bench-analysis} present GRIT-Bench results. GRIT-7B achieves the largest improvement over the base model in both Accuracy ($+11.7$) and Reasoning Fidelity ($+5.9$). This suggests that GRIT not only improves answer correctness but also guides the model to arrive at correct answers through faithful reading of visual information. Across all models, performance generally degrades as text length and reasoning depth increase. The difficulty classification of GRIT-Bench is thus consistent with actual model behavior.

In Figure~\ref{fig:grit-bench-analysis}(a), most models perform on par with or better than the base model in the Short setting, but several begin to fall below the base in Medium, and in the Long setting, nearly all models except GRIT score lower than the base. This suggests that RLVR training can impair reasoning performance on longer text.

In Figure~\ref{fig:grit-bench-analysis}(b), the performance drop due to reasoning depth is relatively small compared to text length. In text-rich reasoning, where models must read and reason over text in images, the volume of textual information within the image is a more critical bottleneck than reasoning complexity.

Notably, Vision-R1a~\citep{vision-r1} improves in Overall Acc ($+2.8$) over the base model but degrades in RF ($-1.3$). Since RF measures the proportion of reasoning steps consistent with the source text and penalizes unsupported claims, this result indicates that while the model answers correctly more often after RL training, it more frequently misreads text in the image or asserts content not present in the source during reasoning. Acc alone cannot detect this pattern, so reasoning faithfulness must be evaluated alongside answer correctness.

\subsection{Ablation Study}
\label{sec:ablation}
\paragraph{Component Contribution.}
\begin{table}[t]
\centering
\caption{Ablation study on component contributions. Each row adds a component to the configuration it is indented under; arrows denote the change from that configuration.}
\label{tab:ablation-components}
\resizebox{1.0\columnwidth}{!}{%
\begin{tabular}{l|c|c}
\toprule
\multirow{-1}{*}{\textbf{Model}} & \textbf{OCR-Reasoning} & \textbf{GRIT-Bench} \\
\midrule
Qwen2.5-VL-7B & 20.58 & 66.0 \\
\raisebox{0.5ex}{$\llcorner$} + GRPO & 18.99 {\scriptsize\textcolor{red}{($\downarrow$ 1.59)}} & 63.9 {\scriptsize\textcolor{red}{($\downarrow$ 2.1)}} \\
\raisebox{0.5ex}{$\llcorner$} + DAPO & 19.55 {\scriptsize\textcolor{red}{($\downarrow$ 1.03)}} & 64.7 {\scriptsize\textcolor{red}{($\downarrow$ 1.3)}} \\
\rowcolor{gray!10}
\quad\raisebox{0.5ex}{$\llcorner$} + Gradient Masking & 24.04 {\scriptsize\textcolor{green!50!black}{($\uparrow$ 4.49)}} & 74.9 {\scriptsize\textcolor{green!50!black}{($\uparrow$ 10.2)}} \\
\rowcolor{gray!20}
\quad\quad\raisebox{0.5ex}{$\llcorner$} + Attention Reward & 25.54 {\scriptsize\textcolor{green!50!black}{($\uparrow$ 1.50)}} & 77.7 {\scriptsize\textcolor{green!50!black}{($\uparrow$ 2.8)}} \\
\bottomrule
\end{tabular}
}
\end{table}

Table~\ref{tab:ablation-components} shows the contribution of each component. GRPO and DAPO both degrade performance relative to the base model on text-rich reasoning. However, adding gradient masking yields the largest single improvement, demonstrating the effectiveness of our approach in addressing the limitations of standard RLVR. Attention reward provides further gains by incorporating response-level grounding quality into the reward signal.

\paragraph{Masking Strategy.}
\begin{table}[t]
\centering
\caption{Ablation on masking strategies (without Attention Reward). Inverted Masking masks the complementary set of tokens selected by GRIT's filtering. Arrows denote the change from DAPO-7B.}
\label{tab:ablation-masking}
\resizebox{1.0\columnwidth}{!}{%
\begin{tabular}{l|c|c}
\toprule
\multirow{-1}{*}{\textbf{Masking Strategy}} & \textbf{OCR-Reasoning} & \textbf{GRIT-Bench} \\
\midrule
DAPO-7B & 19.55 & 64.7 \\
\raisebox{0.5ex}{$\llcorner$} + Inverted Masking & 17.49 {\scriptsize\textcolor{red}{($\downarrow$ 2.06)}} & 59.0 {\scriptsize\textcolor{red}{($\downarrow$ 5.7)}} \\
\raisebox{0.5ex}{$\llcorner$} + Random Masking & 19.46 {\scriptsize\textcolor{red}{($\downarrow$ 0.09)}} & 62.4 {\scriptsize\textcolor{red}{($\downarrow$ 2.3)}} \\
\rowcolor{gray!20}
\textbf{\raisebox{0.5ex}{$\llcorner$} + GRIT Masking (Ours)} & \textbf{24.04} {\scriptsize\textcolor{green!50!black}{($\uparrow$ 4.49)}} & \textbf{74.9} {\scriptsize\textcolor{green!50!black}{($\uparrow$ 10.2)}} \\
\bottomrule
\end{tabular}
}
\end{table}

Table~\ref{tab:ablation-masking} compares masking strategies under the same configuration without attention reward. Inverted masking, which masks well-grounded tokens instead, degrades performance below DAPO, indicating that masking well-grounded tokens actively harms learning. Random masking also slightly degrades performance from DAPO, suggesting that masking tokens without regard to their visual attention patterns provides no benefit. In contrast, GRIT masking achieves substantial gains. The improvement therefore comes from the attention-based selection, not from masking alone.

\paragraph{Effect of Masking Ratio and Head Count.}
\begin{table}[t]
\centering
\caption{Ablation on masking ratio and number of Visual Tracker Heads. Default setting (50/50 ratio, 5 heads) is highlighted.}
\label{tab:ablation-hyperparams}
\resizebox{0.8\columnwidth}{!}{%
\begin{tabular}{cc|c|c}
\toprule
\textbf{Masking Ratio} & \textbf{Heads} & \textbf{OCR-Reasoning} & \textbf{GRIT-Bench} \\
\midrule
30/30 & 5 & 23.57 & 75.1 \\
\rowcolor{gray!20}
\textbf{50/50} & \textbf{5} & \textbf{25.54} & \textbf{77.7} \\
70/70 & 5 & 24.70 & 76.4 \\
\midrule
50/50 & 3 & 25.26 & 77.2 \\
\rowcolor{gray!20}
\textbf{50/50} & \textbf{5} & \textbf{25.54} & \textbf{77.7} \\
50/50 & 10 & 24.51 & 76.7 \\
\bottomrule
\end{tabular}
}
\end{table}

Table~\ref{tab:ablation-hyperparams} examines the effect of masking ratio and the number of Visual Tracker Heads. For masking ratio, 50/50 performs best. For head count, 5 heads yields the best performance, though results are relatively robust across different counts.

\section{Conclusion}

We defined Correct-but-Visually-Misgrounded Reasoning (CVMR), where VLMs arrive at correct answers despite misperceiving visual information, and showed that it occurs frequently in text-rich reasoning with distinct attention patterns. Building on this, we proposed GRIT, a training framework that incorporates visual grounding into RLVR training through attention-based gradient masking and attention reward. We also constructed GRIT-Bench, a text-rich reasoning benchmark that enables systematic evaluation across text length and reasoning depth. While existing RLVR methods mostly degrade or show marginal improvement over the base model on text-rich reasoning, GRIT achieves considerable improvements.

\clearpage

\section{Acknowledgments}
\label{sec:acknowledgement}
This work was supported by the National Research Foundation (NRF) grant funded by the Korea government (MSIT) [RS-2025-00562400] and [RS-2022-NR068754]. The authors acknowledge the Urban Big Data and AI Institute of the University of Seoul supercomputing resources (http://ubai.uos.ac.kr) made available for conducting the research reported in this paper.

{
    \small
    \bibliographystyle{ieeenat_fullname}
    \bibliography{main}

\begin{thebibliography}{42}
\providecommand{\natexlab}[1]{#1}
\providecommand{\url}[1]{\texttt{#1}}
\expandafter\ifx\csname urlstyle\endcsname\relax
  \providecommand{\doi}[1]{doi: #1}\else
  \providecommand{\doi}{doi: \begingroup \urlstyle{rm}\Url}\fi

\bibitem[Baek et~al.(2025)Baek, Chang, Ryu, and Lee]{ocr-heads}
Ingeol Baek, Hwan Chang, Sunghyun Ryu, and Hwanhee Lee.
\newblock How do large vision-language models see text in image? unveiling the distinctive role of ocr heads.
\newblock In \emph{Proceedings of the 2025 Conference on Empirical Methods in Natural Language Processing}, pages 20452--20464, 2025.

\bibitem[Bai et~al.(2025{\natexlab{a}})Bai, Cai, Chen, Chen, Chen, Cheng, Deng, Ding, Gao, Ge, et~al.]{qwen3vl}
Shuai Bai, Yuxuan Cai, Ruizhe Chen, Keqin Chen, Xionghui Chen, Zesen Cheng, Lianghao Deng, Wei Ding, Chang Gao, Chunjiang Ge, et~al.
\newblock Qwen3-vl technical report.
\newblock \emph{arXiv preprint arXiv:2511.21631}, 2025{\natexlab{a}}.

\bibitem[Bai et~al.(2025{\natexlab{b}})Bai, Chen, Liu, Wang, Ge, Song, Dang, Wang, Wang, Tang, Zhong, Zhu, Yang, Li, Wan, Wang, Ding, Fu, Xu, Ye, Zhang, Xie, Cheng, Zhang, Yang, Xu, and Lin]{qwen25vl}
Shuai Bai, Keqin Chen, Xuejing Liu, Jialin Wang, Wenbin Ge, Sibo Song, Kai Dang, Peng Wang, Shijie Wang, Jun Tang, Humen Zhong, Yuanzhi Zhu, Mingkun Yang, Zhaohai Li, Jianqiang Wan, Pengfei Wang, Wei Ding, Zheren Fu, Yiheng Xu, Jiabo Ye, Xi Zhang, Tianbao Xie, Zesen Cheng, Hang Zhang, Zhibo Yang, Haiyang Xu, and Junyang Lin.
\newblock Qwen2.5-vl technical report, 2025{\natexlab{b}}.

\bibitem[Chen et~al.(2025)Chen, Tu, Wang, Liu, Tang, Du, Zhou, and Xie]{vlaa-thinker}
Hardy Chen, Haoqin Tu, Fali Wang, Hui Liu, Xianfeng Tang, Xinya Du, Yuyin Zhou, and Cihang Xie.
\newblock Sft or rl? an early investigation into training r1-like reasoning large vision-language models.
\newblock \emph{arXiv preprint arXiv:2504.11468}, 2025.

\bibitem[Chen et~al.(2024)Chen, Li, Dong, Zhang, Zang, Chen, Duan, Wang, Qiao, Lin, et~al.]{mmstar}
Lin Chen, Jinsong Li, Xiaoyi Dong, Pan Zhang, Yuhang Zang, Zehui Chen, Haodong Duan, Jiaqi Wang, Yu Qiao, Dahua Lin, et~al.
\newblock Are we on the right way for evaluating large vision-language models?
\newblock \emph{Advances in Neural Information Processing Systems}, 37:\penalty0 27056--27087, 2024.

\bibitem[Chen et~al.(2021)Chen, Chen, Smiley, Shah, Borova, Langdon, Moussa, Beane, Huang, Routledge, et~al.]{finqa}
Zhiyu Chen, Wenhu Chen, Charese Smiley, Sameena Shah, Iana Borova, Dylan Langdon, Reema Moussa, Matt Beane, Ting-Hao Huang, Bryan~R Routledge, et~al.
\newblock Finqa: A dataset of numerical reasoning over financial data.
\newblock In \emph{Proceedings of the 2021 Conference on Empirical Methods in Natural Language Processing}, pages 3697--3711, 2021.

\bibitem[Chen et~al.(2022)Chen, Li, Smiley, Ma, Shah, and Wang]{convfinqa}
Zhiyu Chen, Shiyang Li, Charese Smiley, Zhiqiang Ma, Sameena Shah, and William~Yang Wang.
\newblock Convfinqa: Exploring the chain of numerical reasoning in conversational finance question answering.
\newblock In \emph{Proceedings of the 2022 conference on empirical methods in natural language processing}, pages 6279--6292, 2022.

\bibitem[Cheng et~al.(2022)Cheng, Dong, Wang, Jia, Guo, Gao, Han, Lou, and Zhang]{hitab}
Zhoujun Cheng, Haoyu Dong, Zhiruo Wang, Ran Jia, Jiaqi Guo, Yan Gao, Shi Han, Jian-Guang Lou, and Dongmei Zhang.
\newblock Hitab: A hierarchical table dataset for question answering and natural language generation.
\newblock In \emph{Proceedings of the 60th Annual Meeting of the Association for Computational Linguistics (Volume 1: Long Papers)}, pages 1094--1110, 2022.

\bibitem[Dua et~al.(2019)Dua, Wang, Dasigi, Stanovsky, Singh, and Gardner]{drop}
Dheeru Dua, Yizhong Wang, Pradeep Dasigi, Gabriel Stanovsky, Sameer Singh, and Matt Gardner.
\newblock Drop: A reading comprehension benchmark requiring discrete reasoning over paragraphs.
\newblock In \emph{Proceedings of the 2019 Conference of the North American Chapter of the Association for Computational Linguistics: Human Language Technologies, Volume 1 (Long and Short Papers)}, pages 2368--2378, 2019.

\bibitem[Duan et~al.(2024)Duan, Yang, Qiao, Fang, Chen, Liu, Dong, Zang, Zhang, Wang, et~al.]{vlmevalkit}
Haodong Duan, Junming Yang, Yuxuan Qiao, Xinyu Fang, Lin Chen, Yuan Liu, Xiaoyi Dong, Yuhang Zang, Pan Zhang, Jiaqi Wang, et~al.
\newblock Vlmevalkit: An open-source toolkit for evaluating large multi-modality models.
\newblock In \emph{Proceedings of the 32nd ACM International Conference on Multimedia}, pages 11198--11201, 2024.

\bibitem[Guan et~al.(2024)Guan, Liu, Wu, Xian, Li, Liu, Wang, Chen, Huang, Yacoob, et~al.]{hallusionbench}
Tianrui Guan, Fuxiao Liu, Xiyang Wu, Ruiqi Xian, Zongxia Li, Xiaoyu Liu, Xijun Wang, Lichang Chen, Furong Huang, Yaser Yacoob, et~al.
\newblock Hallusionbench: an advanced diagnostic suite for entangled language hallucination and visual illusion in large vision-language models.
\newblock In \emph{Proceedings of the IEEE/CVF conference on computer vision and pattern recognition}, pages 14375--14385, 2024.

\bibitem[Guo et~al.(2025)Guo, Yang, Zhang, Song, Wang, Zhu, Xu, Zhang, Ma, Bi, et~al.]{deepseek-r1}
Daya Guo, Dejian Yang, Haowei Zhang, Junxiao Song, Peiyi Wang, Qihao Zhu, Runxin Xu, Ruoyu Zhang, Shirong Ma, Xiao Bi, et~al.
\newblock Deepseek-r1: Incentivizing reasoning capability in llms via reinforcement learning.
\newblock \emph{arXiv preprint arXiv:2501.12948}, 2025.

\bibitem[Huang et~al.(2025{\natexlab{a}})Huang, Shi, Peng, Lai, Xie, and Jin]{ocr-reasoning}
Mingxin Huang, Yongxin Shi, Dezhi Peng, Songxuan Lai, Zecheng Xie, and Lianwen Jin.
\newblock Ocr-reasoning benchmark: Unveiling the true capabilities of mllms in complex text-rich image reasoning.
\newblock \emph{arXiv preprint arXiv:2505.17163}, 2025{\natexlab{a}}.

\bibitem[Huang et~al.(2025{\natexlab{b}})Huang, Jia, Zhai, Cao, Ye, Zhao, Xu, Tang, Hu, and Lin]{vision-r1}
Wenxuan Huang, Bohan Jia, Zijie Zhai, Shaosheng Cao, Zheyu Ye, Fei Zhao, Zhe Xu, Xu Tang, Yao Hu, and Shaohui Lin.
\newblock Vision-r1: Incentivizing reasoning capability in multimodal large language models.
\newblock \emph{arXiv preprint arXiv:2503.06749}, 2025{\natexlab{b}}.

\bibitem[Kang et~al.(2025)Kang, Kim, Kim, and Hwang]{few-attn-heads}
Seil Kang, Jinyeong Kim, Junhyeok Kim, and Seong~Jae Hwang.
\newblock Your large vision-language model only needs a few attention heads for visual grounding.
\newblock In \emph{Proceedings of the Computer Vision and Pattern Recognition Conference}, pages 9339--9350, 2025.

\bibitem[Li et~al.(2025)Li, Yu, Huang, Liu, Liang, Liu, Che, Yu, Boyd-Graber, Mi, et~al.]{vision-sr1}
Zongxia Li, Wenhao Yu, Chengsong Huang, Rui Liu, Zhenwen Liang, Fuxiao Liu, Jingxi Che, Dian Yu, Jordan Boyd-Graber, Haitao Mi, et~al.
\newblock Self-rewarding vision-language model via reasoning decomposition.
\newblock \emph{arXiv preprint arXiv:2508.19652}, 2025.

\bibitem[Liu et~al.(2024)Liu, Xue, Chen, Chen, Zhao, Wang, Hou, Li, and Peng]{hallucination-survey}
Hanchao Liu, Wenyuan Xue, Yifei Chen, Dapeng Chen, Xiutian Zhao, Ke Wang, Liping Hou, Rongjun Li, and Wei Peng.
\newblock A survey on hallucination in large vision-language models.
\newblock \emph{arXiv preprint arXiv:2402.00253}, 2024.

\bibitem[Lu et~al.(2022)Lu, Mishra, Xia, Qiu, Chang, Zhu, Tafjord, Clark, and Kalyan]{scienceqa}
Pan Lu, Swaroop Mishra, Tanglin Xia, Liang Qiu, Kai-Wei Chang, Song-Chun Zhu, Oyvind Tafjord, Peter Clark, and Ashwin Kalyan.
\newblock Learn to explain: Multimodal reasoning via thought chains for science question answering.
\newblock \emph{Advances in neural information processing systems}, 35:\penalty0 2507--2521, 2022.

\bibitem[Masry et~al.(2022)Masry, Do, Tan, Joty, and Hoque]{chartqa}
Ahmed Masry, Xuan~Long Do, Jia~Qing Tan, Shafiq Joty, and Enamul Hoque.
\newblock Chartqa: A benchmark for question answering about charts with visual and logical reasoning.
\newblock In \emph{Findings of the association for computational linguistics: ACL 2022}, pages 2263--2279, 2022.

\bibitem[Masry et~al.(2025)Masry, Islam, Ahmed, Bajaj, Kabir, Kartha, Laskar, Rahman, Rahman, Shahmohammadi, et~al.]{chartqapro}
Ahmed Masry, Mohammed~Saidul Islam, Mahir Ahmed, Aayush Bajaj, Firoz Kabir, Aaryaman Kartha, Md~Tahmid~Rahman Laskar, Mizanur Rahman, Shadikur Rahman, Mehrad Shahmohammadi, et~al.
\newblock Chartqapro: A more diverse and challenging benchmark for chart question answering.
\newblock In \emph{Findings of the Association for Computational Linguistics: ACL 2025}, pages 19123--19151, 2025.

\bibitem[Mathew et~al.(2021)Mathew, Karatzas, and Jawahar]{docvqa}
Minesh Mathew, Dimosthenis Karatzas, and CV Jawahar.
\newblock Docvqa: A dataset for vqa on document images.
\newblock In \emph{Proceedings of the IEEE/CVF winter conference on applications of computer vision}, pages 2200--2209, 2021.

\bibitem[Meng et~al.(2025)Meng, Du, Liu, Zhou, Lu, Fu, Han, Shi, Wang, He, et~al.]{mm-eureka}
Fanqing Meng, Lingxiao Du, Zongkai Liu, Zhixiang Zhou, Quanfeng Lu, Daocheng Fu, Tiancheng Han, Botian Shi, Wenhai Wang, Junjun He, et~al.
\newblock Mm-eureka: Exploring the frontiers of multimodal reasoning with rule-based reinforcement learning.
\newblock \emph{arXiv preprint arXiv:2503.07365}, 2025.

\bibitem[Pasupat and Liang(2015)]{wtq}
Panupong Pasupat and Percy Liang.
\newblock Compositional semantic parsing on semi-structured tables.
\newblock In \emph{Proceedings of the 53rd Annual Meeting of the Association for Computational Linguistics and the 7th International Joint Conference on Natural Language Processing (Volume 1: Long Papers)}, pages 1470--1480, 2015.

\bibitem[Peng et~al.(2025)Peng, Wang, Wang, Wei, Pei, Qiu, Jian, Hao, Pan, Xie, et~al.]{skywork-r1v}
Yi Peng, Peiyu Wang, Xiaokun Wang, Yichen Wei, Jiangbo Pei, Weijie Qiu, Ai Jian, Yunzhuo Hao, Jiachun Pan, Tianyidan Xie, et~al.
\newblock Skywork r1v: Pioneering multimodal reasoning with chain-of-thought.
\newblock \emph{arXiv preprint arXiv:2504.05599}, 2025.

\bibitem[Schwenk et~al.(2022)Schwenk, Khandelwal, Clark, Marino, and Mottaghi]{a-okvqa}
Dustin Schwenk, Apoorv Khandelwal, Christopher Clark, Kenneth Marino, and Roozbeh Mottaghi.
\newblock A-okvqa: A benchmark for visual question answering using world knowledge.
\newblock In \emph{European conference on computer vision}, pages 146--162. Springer, 2022.

\bibitem[Shao et~al.(2024)Shao, Wang, Zhu, Xu, Song, Bi, Zhang, Zhang, Li, Wu, et~al.]{deepseekmath}
Zhihong Shao, Peiyi Wang, Qihao Zhu, Runxin Xu, Junxiao Song, Xiao Bi, Haowei Zhang, Mingchuan Zhang, YK Li, Yang Wu, et~al.
\newblock Deepseekmath: Pushing the limits of mathematical reasoning in open language models.
\newblock \emph{arXiv preprint arXiv:2402.03300}, 2024.

\bibitem[Singh et~al.(2019)Singh, Natarajan, Shah, Jiang, Chen, Batra, Parikh, and Rohrbach]{textvqa}
Amanpreet Singh, Vivek Natarajan, Meet Shah, Yu Jiang, Xinlei Chen, Dhruv Batra, Devi Parikh, and Marcus Rohrbach.
\newblock Towards vqa models that can read.
\newblock In \emph{Proceedings of the IEEE/CVF conference on computer vision and pattern recognition}, pages 8317--8326, 2019.

\bibitem[Singh et~al.(2025)Singh, Fry, Perelman, Tart, Ganesh, El-Kishky, McLaughlin, Low, Ostrow, Ananthram, et~al.]{gpt-5}
Aaditya Singh, Adam Fry, Adam Perelman, Adam Tart, Adi Ganesh, Ahmed El-Kishky, Aidan McLaughlin, Aiden Low, AJ Ostrow, Akhila Ananthram, et~al.
\newblock Openai gpt-5 system card.
\newblock \emph{arXiv preprint arXiv:2601.03267}, 2025.

\bibitem[Team et~al.(2025)Team, Du, Gao, Xing, Jiang, Chen, Li, Xiao, Du, Liao, et~al.]{kimi-k15}
Kimi Team, Angang Du, Bofei Gao, Bowei Xing, Changjiu Jiang, Cheng Chen, Cheng Li, Chenjun Xiao, Chenzhuang Du, Chonghua Liao, et~al.
\newblock Kimi k1. 5: Scaling reinforcement learning with llms.
\newblock \emph{arXiv preprint arXiv:2501.12599}, 2025.

\bibitem[Voita et~al.(2019)Voita, Talbot, Moiseev, Sennrich, and Titov]{voita2019heads}
Elena Voita, David Talbot, Fedor Moiseev, Rico Sennrich, and Ivan Titov.
\newblock Analyzing multi-head self-attention: Specialized heads do the heavy lifting, the rest can be pruned.
\newblock In \emph{Proceedings of the 57th annual meeting of the association for computational linguistics}, pages 5797--5808, 2019.

\bibitem[Wadhawan et~al.(2024)Wadhawan, Bansal, Chang, and Peng]{contextual}
Rohan Wadhawan, Hritik Bansal, Kai-Wei Chang, and Nanyun Peng.
\newblock Contextual: Evaluating context-sensitive text-rich visual reasoning in large multimodal models.
\newblock \emph{arXiv preprint arXiv:2401.13311}, 2024.

\bibitem[Wang et~al.(2025)Wang, Qu, Huang, Chu, Lin, and Chen]{vl-rethinker}
Haozhe Wang, Chao Qu, Zuming Huang, Wei Chu, Fangzhen Lin, and Wenhu Chen.
\newblock Vl-rethinker: Incentivizing self-reflection of vision-language models with reinforcement learning.
\newblock \emph{arXiv preprint arXiv:2504.08837}, 2025.

\bibitem[Wang et~al.(2024)Wang, Xia, He, Chen, Liu, Zhu, Liang, Wu, Liu, Malladi, et~al.]{charxiv}
Zirui Wang, Mengzhou Xia, Luxi He, Howard Chen, Yitao Liu, Richard Zhu, Kaiqu Liang, Xindi Wu, Haotian Liu, Sadhika Malladi, et~al.
\newblock Charxiv: Charting gaps in realistic chart understanding in multimodal llms.
\newblock \emph{Advances in Neural Information Processing Systems}, 37:\penalty0 113569--113697, 2024.

\bibitem[{xAI}(2024)]{realworldqa}
{xAI}.
\newblock Realworldqa.
\newblock \url{https://huggingface.co/datasets/xai-org/RealworldQA}, 2024.

\bibitem[Yang et~al.(2025)Yang, He, Pan, Jiang, Deng, Yang, Lu, Yin, Rao, Zhu, et~al.]{r1-onevision}
Yi Yang, Xiaoxuan He, Hongkun Pan, Xiyan Jiang, Yan Deng, Xingtao Yang, Haoyu Lu, Dacheng Yin, Fengyun Rao, Minfeng Zhu, et~al.
\newblock R1-onevision: Advancing generalized multimodal reasoning through cross-modal formalization.
\newblock In \emph{Proceedings of the IEEE/CVF International Conference on Computer Vision}, pages 2376--2385, 2025.

\bibitem[Yu et~al.(2026)Yu, Zhang, Zhu, Yuan, Zuo, Yue, Dai, Fan, Liu, Liu, et~al.]{dapo}
Qiying Yu, Zheng Zhang, Ruofei Zhu, Yufeng Yuan, Xiaochen Zuo, Yu Yue, Weinan Dai, Tiantian Fan, Gaohong Liu, Lingjun Liu, et~al.
\newblock Dapo: An open-source llm reinforcement learning system at scale.
\newblock \emph{Advances in Neural Information Processing Systems}, 38:\penalty0 113222--113244, 2026.

\bibitem[Yu et~al.(2025)Yu, Yang, Liu, and Bai]{docthinker}
Wenwen Yu, Zhibo Yang, Yuliang Liu, and Xiang Bai.
\newblock Docthinker: Explainable multimodal large language models with rule-based reinforcement learning for document understanding.
\newblock In \emph{Proceedings of the IEEE/CVF International Conference on Computer Vision}, pages 837--847, 2025.

\bibitem[Zhan et~al.(2025)Zhan, Zhu, Zheng, Zhao, Yang, Tang, and Wang]{vision-r1-star}
Yufei Zhan, Yousong Zhu, Shurong Zheng, Hongyin Zhao, Fan Yang, Ming Tang, and Jinqiao Wang.
\newblock Vision-r1: Evolving human-free alignment in large vision-language models via vision-guided reinforcement learning.
\newblock \emph{arXiv preprint arXiv:2503.18013}, 2025.

\bibitem[Zhao et~al.(2022)Zhao, Li, Li, and Zhang]{multihiertt}
Yilun Zhao, Yunxiang Li, Chenying Li, and Rui Zhang.
\newblock Multihiertt: Numerical reasoning over multi hierarchical tabular and textual data.
\newblock In \emph{Proceedings of the 60th Annual Meeting of the Association for Computational Linguistics (Volume 1: Long Papers)}, pages 6588--6600, 2022.

\bibitem[Zheng et~al.(2025)Zheng, Qian, Tang, and Yang]{why-lvlm-hallucinate-longer}
Ge Zheng, Jiaye Qian, Jiajin Tang, and Sibei Yang.
\newblock Why lvlms are more prone to hallucinations in longer responses: The role of context.
\newblock In \emph{Proceedings of the IEEE/CVF International Conference on Computer Vision}, pages 4101--4113, 2025.

\bibitem[Zheng et~al.(2024)Zheng, Wang, Huang, Song, Yang, Tang, Xiong, and Li]{attn-heads-survey}
Zifan Zheng, Yezhaohui Wang, Yuxin Huang, Shichao Song, Mingchuan Yang, Bo Tang, Feiyu Xiong, and Zhiyu Li.
\newblock Attention heads of large language models: A survey.
\newblock \emph{arXiv preprint arXiv:2409.03752}, 2024.

\bibitem[Zhu et~al.(2021)Zhu, Lei, Huang, Wang, Zhang, Lv, Feng, and Chua]{tatqa}
Fengbin Zhu, Wenqiang Lei, Youcheng Huang, Chao Wang, Shuo Zhang, Jiancheng Lv, Fuli Feng, and Tat-Seng Chua.
\newblock Tat-qa: A question answering benchmark on a hybrid of tabular and textual content in finance.
\newblock In \emph{Proceedings of the 59th annual meeting of the Association for Computational Linguistics and the 11th international joint conference on natural language processing (volume 1: long papers)}, pages 3277--3287, 2021.

\end{thebibliography}
}

\clearpage
\appendix

\section{Implementation Details}
\label{app:implementation}

\paragraph{Training.}
We use AdamW optimizer with a learning rate of $1 \times 10^{-6}$, cosine learning rate schedule without warmup, and weight decay of $1 \times 10^{-2}$. Training is conducted in bf16 mixed precision with full fine-tuning. We train on the ViRL 39K dataset~\citep{vl-rethinker} for 2 epochs with an effective batch size of 128 prompts per update. GRPO and DAPO baselines are trained under the same configuration. Benchmark evaluation is conducted using VLMEvalKit~\citep{vlmevalkit}.

\paragraph{GRPO Configuration.}
We sample $G = 16$ rollouts per prompt. We remove the KL penalty, apply asymmetric clipping ($\epsilon_l = 0.2$, $\epsilon_h = 0.28$)~\citep{dapo}, and adopt token-level loss normalization. Rollouts are generated using vLLM with temperature 1.0 and top-$p$ 0.99.

\paragraph{GRIT Configuration.}
For gradient masking, tokens in the bottom 50\% of visual attention mass and top 50\% of visual attention entropy within each response are masked. Answer format tags (\texttt{<answer>}, \texttt{<think>}), mathematical operators (\texttt{+\,-\,=\,*\,/}), and punctuation (\texttt{.\,,\,(\,)\,"\,'}) are designated as protected tokens and always retain their gradients. For the final reward, correctness, format, and attention rewards are weighted equally ($w = 1.0$ each).

\paragraph{Prompt Template.}
Each training sample is formatted by appending the following instruction to the question: \textit{``Solve the problem through step-by-step reasoning. Put your reasoning inside \texttt{<think>}...\texttt{</think>} tags, then your final answer inside \texttt{<answer>}...\texttt{</answer>} tags.''}

\section{Visual Tracker Heads}
\label{app:vth}

For Qwen2.5-VL-7B, which has 784 attention heads (28 layers $\times$ 28 heads per layer), we first retain the top 50 heads by mean visual attention mass, then rank them by token-conditioned spatial diversity (Equation~\ref{eq:diversity}) and select the top 5. The selected Visual Tracker Heads are located at (layer, head): (14, 0), (22, 4), (8, 27), (26, 25), and (19, 20), distributed across layers 8, 14, 19, 22, and 26. Figures~\ref{fig:appendix_cvmr_example}, \ref{fig:appendix_cvmr_example2}, and \ref{fig:appendix_cvmr_example3} visualize the attention patterns of these heads, showing that attention shifts to relevant yet different image regions depending on which token is being generated.

\section{Efficient Attention Weight Extraction}
\label{sec:efficiency}

GRIT requires attention weights only during training. Token-level masking itself is a negligible Boolean operation on the policy loss; the measured overhead comes from materializing attention weights, which prevents the selected layers from using the default FlashAttention-style implementation. A naive implementation extracts attention weights from every layer. However, the five Visual Tracker Heads used by GRIT are distributed across only five of the 28 layers in Qwen2.5-VL-7B. We therefore extract attention weights only from these five layers and retain the default efficient attention implementation in all remaining layers.

Table~\ref{tab:training_cost} reports wall-clock measurements under the same Qwen2.5-VL-7B training configuration using three H200 GPUs with 141\,GB of memory each. Selective extraction reduces the relative training cost from $2.52\times$ for naive all-layer extraction to $1.29\times$. This is a one-time training cost: the trained model has no architectural changes and incurs no additional inference-time computation.

\begin{table}[t]
    \centering
    \caption{Wall-clock training cost on Qwen2.5-VL-7B using three H200 GPUs. All settings use the same training configuration, and relative cost is normalized by GRPO.}
    \label{tab:training_cost}
    \resizebox{\columnwidth}{!}{%
    \begin{tabular}{lccc}
        \toprule
        \textbf{Setting} & \textbf{Time/step} & \textbf{Total wall-clock} & \textbf{Relative} \\
        \midrule
        GRPO (no attention extraction)
        & 650 s & 110.1 h & $1.00\times$ \\

        GRIT (all-layer extraction)
        & 1,638 s & 277.6 h & $2.52\times$ \\

        \rowcolor{gray!20}
        \textbf{GRIT (selective extraction, adopted)}
        & \textbf{838 s} & \textbf{142.1 h} & \textbf{$1.29\times$} \\
        \bottomrule
    \end{tabular}%
    }
\end{table}

Further reductions may be possible with kernels that return attention weights only for a small set of requested heads without materializing full attention maps.

\section{GRIT-Bench Construction Details}
\label{app:grit-bench}

GRIT-Bench is constructed from seven publicly available text-based QA datasets. Table~\ref{tab:grit-bench-dataset} shows the number of samples from each source dataset. Source texts are rendered into images using randomized table and document styling, and key reasoning steps are extracted by GPT-5 from the source text, followed by human verification.

\begin{table}[h]
\centering
\caption{Source datasets and sample counts for GRIT-Bench.}
\label{tab:grit-bench-dataset}
\resizebox{0.6\columnwidth}{!}{%
\begin{tabular}{l|c}
\toprule
\textbf{Dataset} & \textbf{Samples} \\
\midrule
ConvFinQA~\citep{convfinqa} & 476 \\
WTQ~\citep{wtq} & 460 \\
DROP~\citep{drop} & 450 \\
TAT-QA~\citep{tatqa} & 428 \\
HiTab~\citep{hitab} & 417 \\
FinQA~\citep{finqa} & 349 \\
MultiHiertt~\citep{multihiertt} & 275 \\
\midrule
Total & 2,855 \\
\bottomrule
\end{tabular}
}
\end{table}

During evaluation, each model receives the rendered image along with the question and the following instruction: \textit{``Solve the problem through step-by-step reasoning. Put your reasoning inside \texttt{<think>}...\texttt{</think>} tags, then your final answer inside \texttt{<answer>}...\texttt{</answer>} tags.''}

For Reasoning Fidelity scoring, GPT-5 mini receives the original source text (not the image), the question, the ground-truth answer, the key reasoning steps, and the model's response. The scoring prompt is as follows:

\begin{quote}
\small
\textit{For each key reasoning step listed above:}\\
\textit{1. Did the model attempt this step in its chain-of-thought? (yes/no)}\\
\textit{2. If attempted, are the facts/numbers cited consistent with the source text? (yes/no)}\\
\textit{--- ``faithful'' means the data the model uses matches the source text. If the model uses wrong numbers or cites non-existent data, it is NOT faithful.}\\[4pt]
\textit{Also judge: Is the model's final answer correct?}\\[4pt]
\textit{Additionally, count how many claims in the model's response contradict or fabricate facts not present in the source text, EXCLUDING the key steps already evaluated above. These are ``ungrounded claims''.}
\end{quote}

\begin{figure*}[t]
\centering
\includegraphics[width=1.0\linewidth]{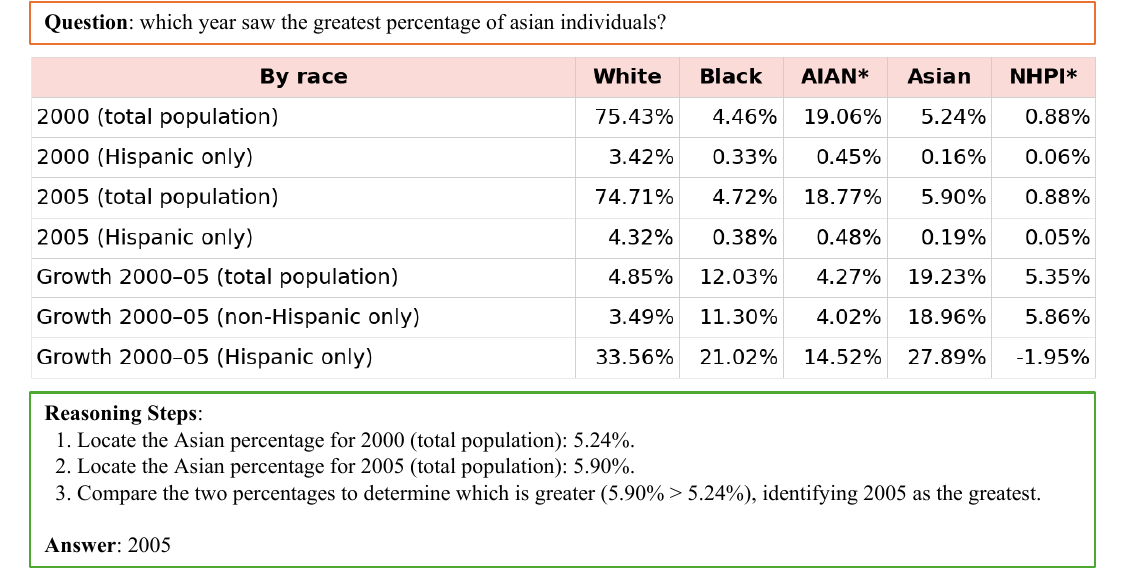}
\caption{GRIT-Bench example with rendered image, question, key reasoning steps, and answer.}
\label{fig:grit-bench-ex1}
\end{figure*}

\begin{figure*}[t]
\centering
\includegraphics[width=1.0\linewidth]{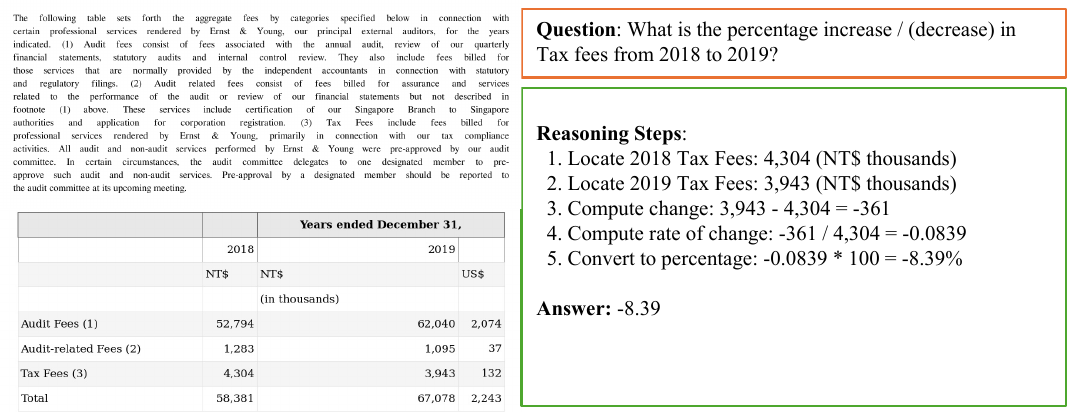}
\caption{GRIT-Bench example with rendered image, question, key reasoning steps, and answer.}
\label{fig:grit-bench-ex2}
\end{figure*}

\begin{figure*}[t]
\centering
\includegraphics[width=1.0\linewidth]{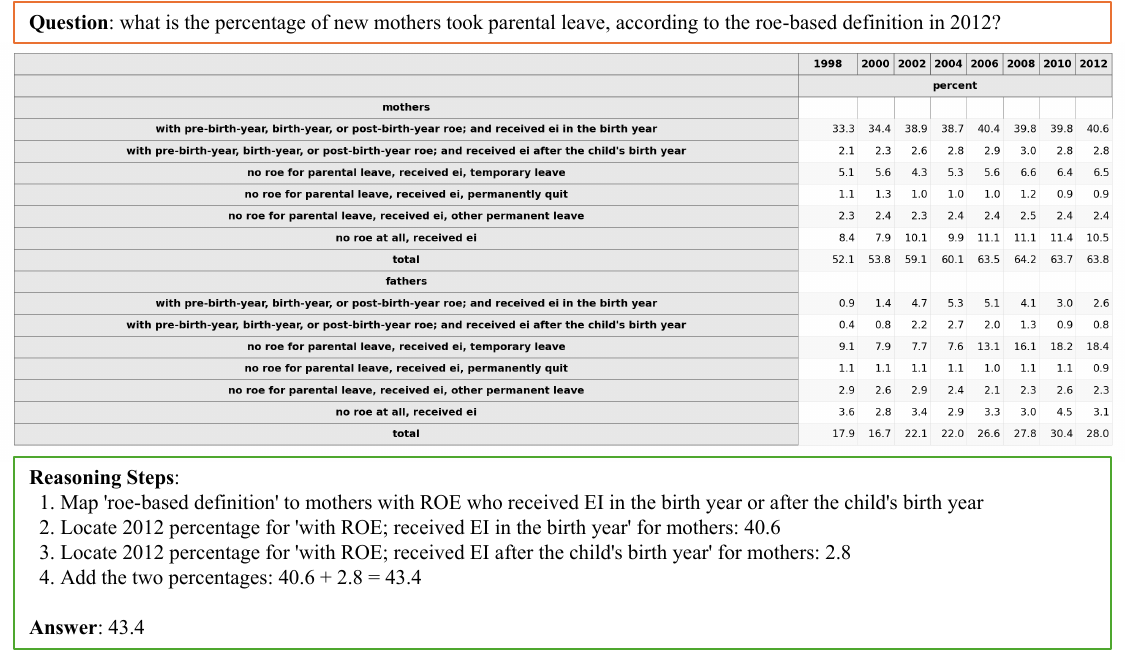}
\caption{GRIT-Bench example with rendered image, question, key reasoning steps, and answer.}
\label{fig:grit-bench-ex3}
\end{figure*}

\begin{figure*}[t]
\centering
\includegraphics[width=1.0\linewidth]{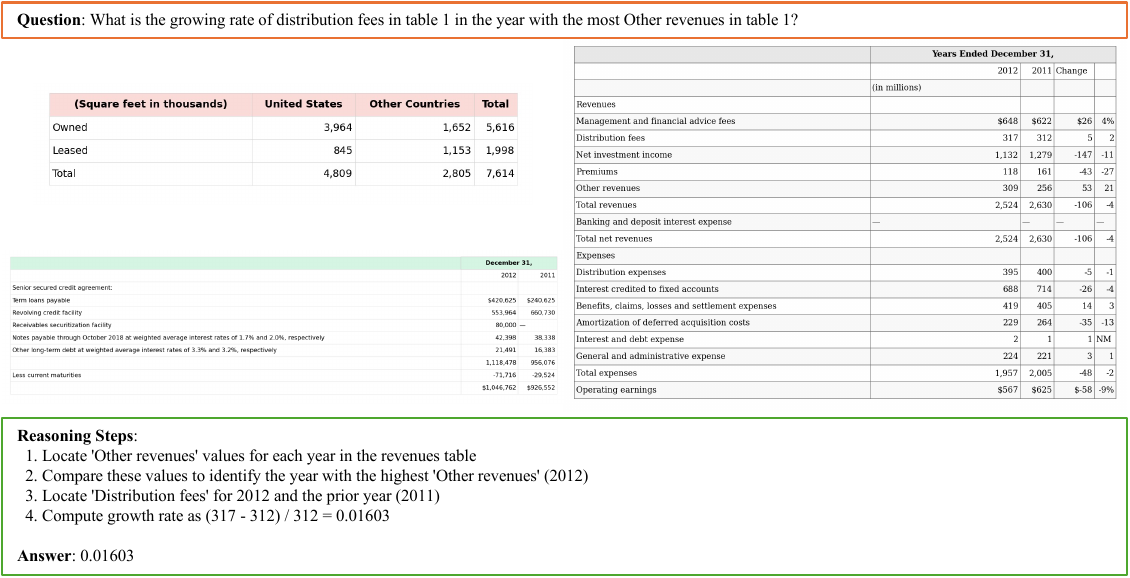}
\caption{GRIT-Bench example with rendered image, question, key reasoning steps, and answer.}
\label{fig:grit-bench-ex4}
\end{figure*}

Figures~\ref{fig:grit-bench-ex1}, \ref{fig:grit-bench-ex2}, \ref{fig:grit-bench-ex3}, and \ref{fig:grit-bench-ex4} show examples from GRIT-Bench, each with the rendered image, question, key reasoning steps, and answer.

\section{Additional CVMR Examples}
\label{app:qualitative}

\begin{figure*}[t]
\centering
\includegraphics[width=0.8\linewidth]{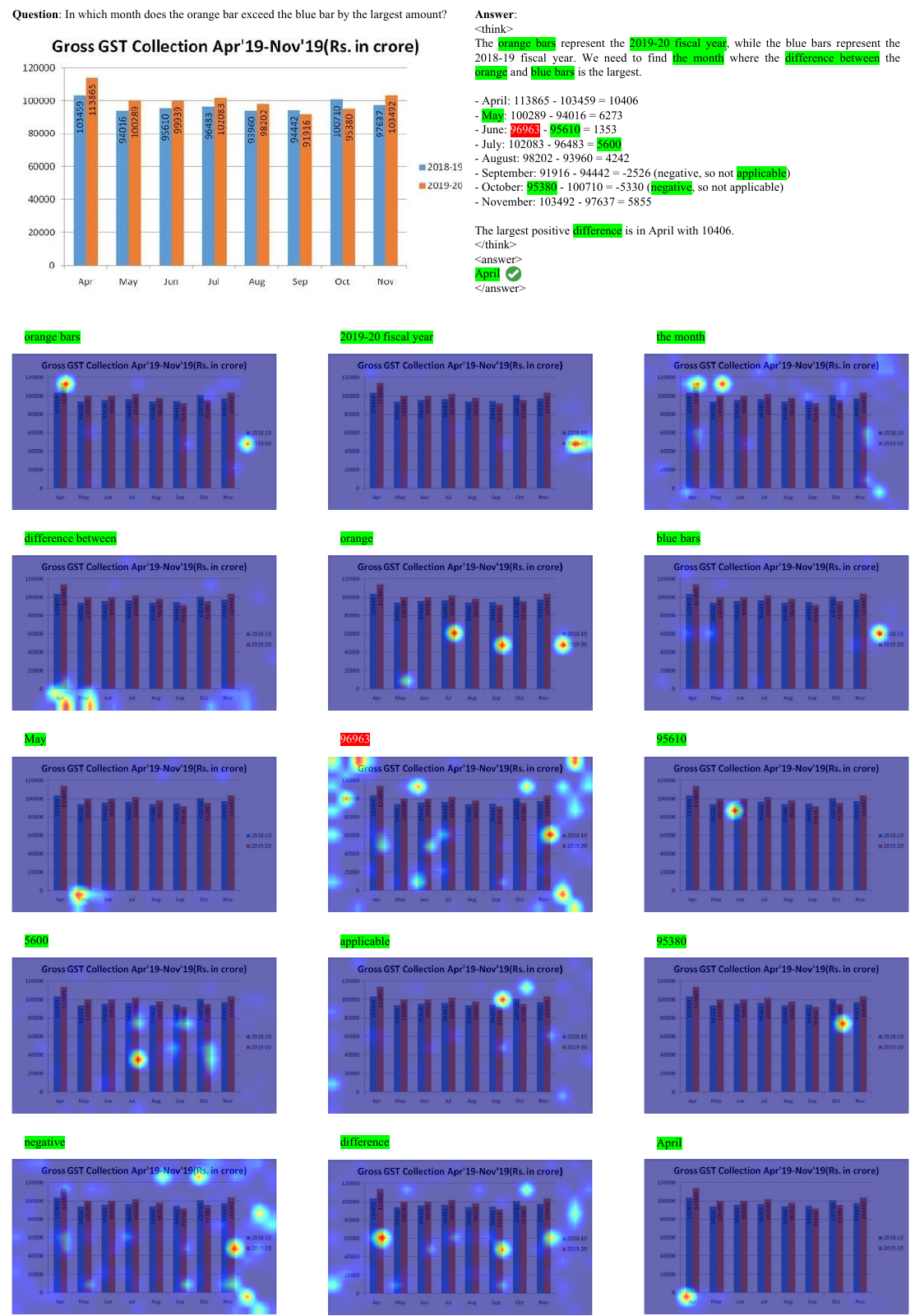}
\caption{Visual Tracker Head attention maps on a CVMR example. Each heatmap corresponds to the highlighted token. \colorbox{red!30}{Red} highlights indicate CVMR tokens; \colorbox{green!30}{green} highlights indicate normal tokens.}
\label{fig:appendix_cvmr_example}
\end{figure*}

\begin{figure*}[t]
\centering
\includegraphics[width=0.8\linewidth]{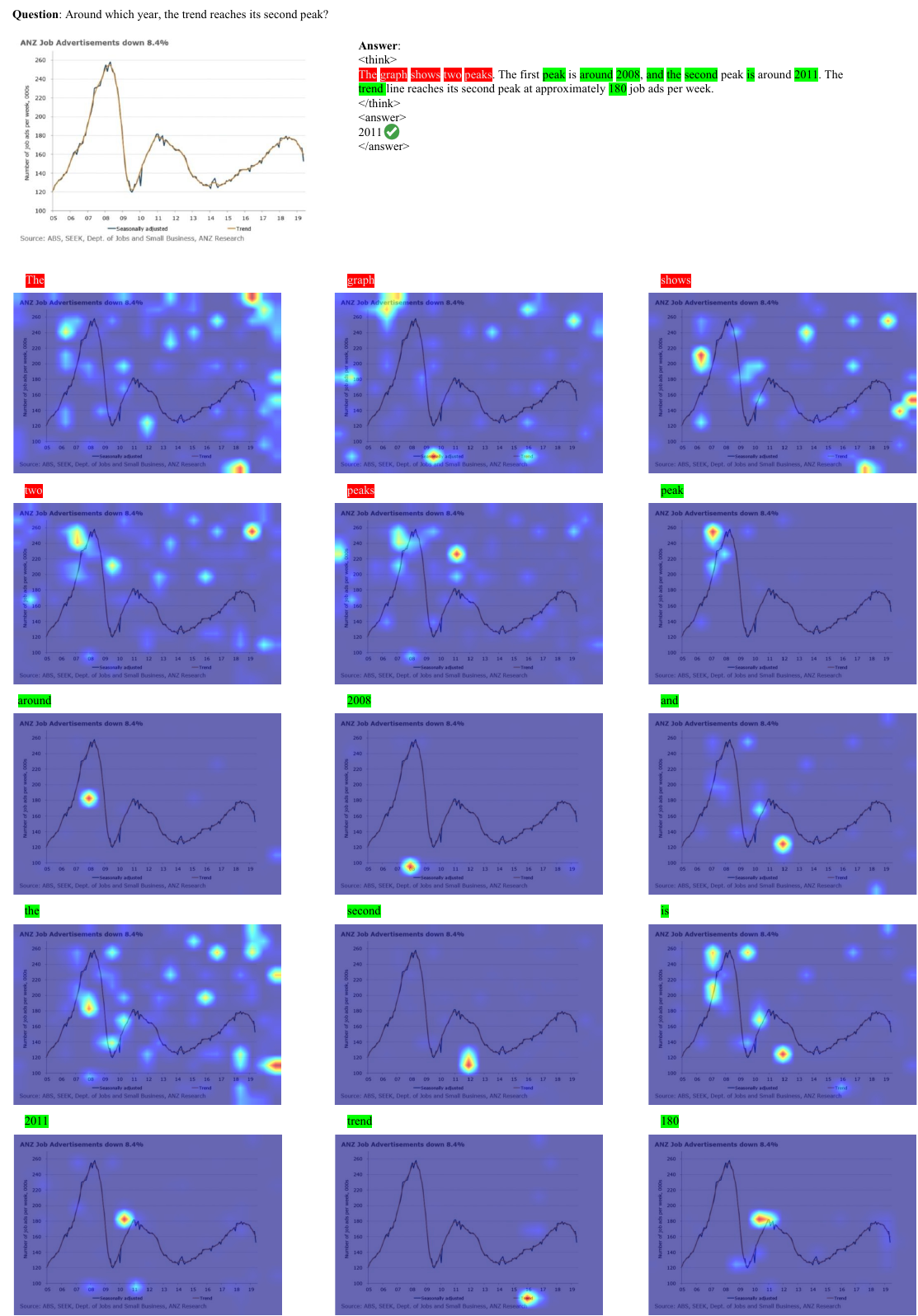}
\caption{Visual Tracker Head attention maps on a CVMR example. Each heatmap corresponds to the highlighted token. \colorbox{red!30}{Red} highlights indicate CVMR tokens; \colorbox{green!30}{green} highlights indicate normal tokens.}
\label{fig:appendix_cvmr_example2}
\end{figure*}

\begin{figure*}[t]
\centering
\includegraphics[width=0.8\linewidth]{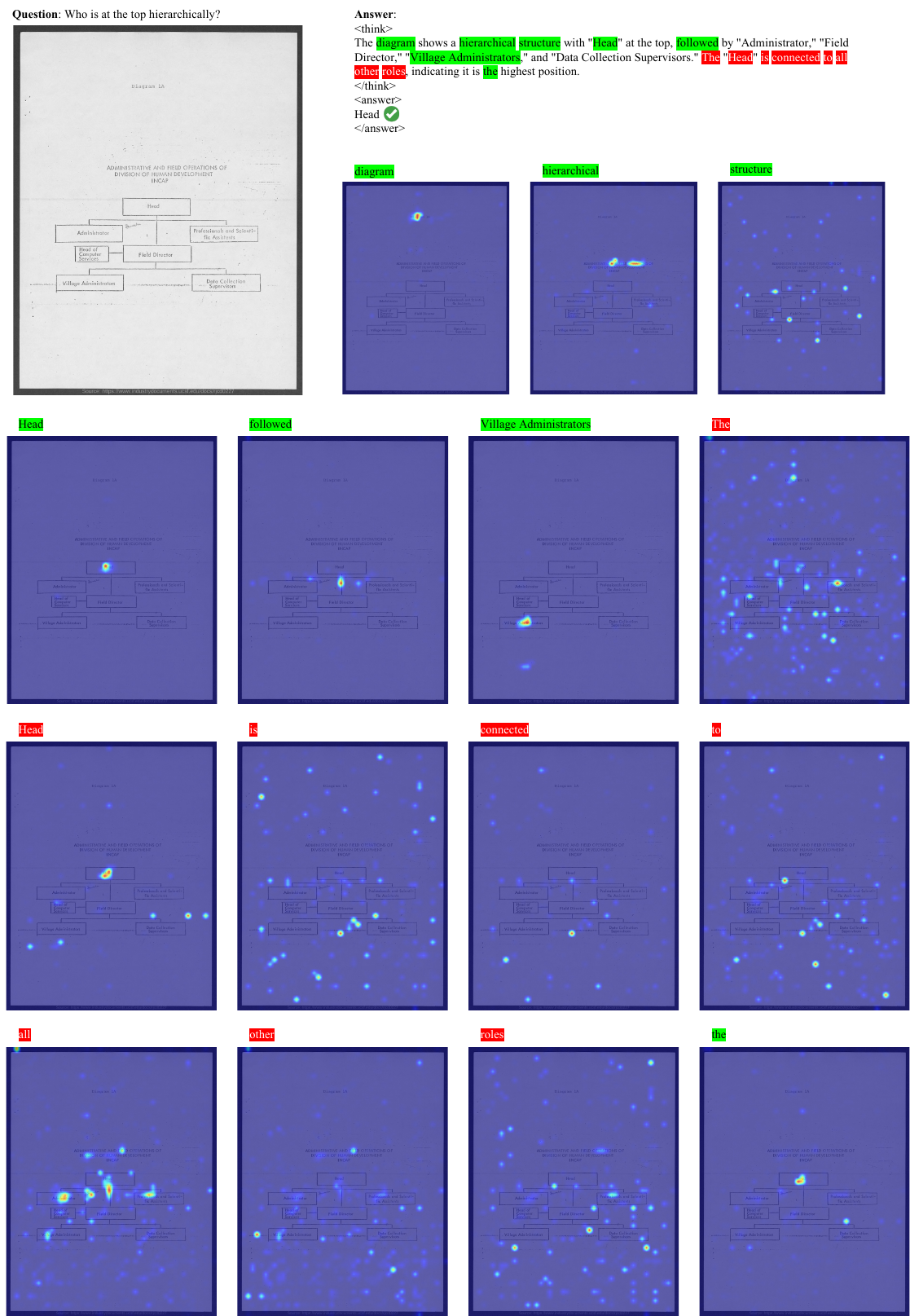}
\caption{Visual Tracker Head attention maps on a CVMR example. Each heatmap corresponds to the highlighted token. \colorbox{red!30}{Red} highlights indicate CVMR tokens; \colorbox{green!30}{green} highlights indicate normal tokens.}
\label{fig:appendix_cvmr_example3}
\end{figure*}

Figures~\ref{fig:appendix_cvmr_example}, \ref{fig:appendix_cvmr_example2}, and \ref{fig:appendix_cvmr_example3} present additional CVMR examples with per-token attention maps from the Visual Tracker Heads. In each example, the model arrives at the correct answer despite misreading parts of the visual information. The attention maps show that normal tokens (green) tend to focus on specific image regions relevant to the token being generated, while CVMR tokens (red) often exhibit more diffuse attention. We note that some normal tokens also display diffuse attention patterns; this is expected, as not every token in a response directly references visual content.

\section{Discussion and Future Work}
\label{app:discussion}

\paragraph{Approximate token filtering.}
GRIT uses visual attention mass and entropy as a lightweight proxy for token-level grounding. The resulting mask is conservative: selected tokens are excluded from the current policy update rather than optimized in the opposite direction. Tables~\ref{tab:ablation-masking} and~\ref{tab:ablation-hyperparams} show that the gains depend on attention-based token selection and remain stable across different masking thresholds and numbers of Visual Tracker Heads. Nevertheless, attention statistics do not perfectly determine whether an individual token is visually grounded. Future work could replace the fixed percentile rule with a calibrated or learned grounding-confidence estimator.

\paragraph{Training efficiency.}
The additional training cost arises from materializing attention weights, rather than from applying the token mask itself. By extracting attention only from the five layers containing Visual Tracker Heads, GRIT reduces the relative wall-clock cost from $2.52\times$ for all-layer extraction to $1.29\times$, while preserving the original model architecture and inference cost. More selective attention kernels could further reduce this overhead.

\paragraph{Scope of GRIT-Bench.}
GRIT-Bench uses rendered inputs to preserve exact source--image alignment, enabling source-aligned RF evaluation and controlled analysis by text length and reasoning depth. This controlled setting does not capture the full variability of real-world layouts, typography, and capture noise. We therefore use GRIT-Bench together with independently constructed text-rich benchmarks, including OCR-Reasoning, ChartQA-PRO, and CharXiv. Future work could extend RF-style evaluation to real documents with verified source text and step-level references.

\section{Transfer to Qwen3-VL-4B}
\label{sec:qwen3}
To evaluate transfer across backbones, we apply GRIT to Qwen3-VL-4B-Instruct~\cite{qwen3vl} using the same training configuration as in the main experiments.
Qwen3-VL departs from Qwen2.5-VL in its core multimodal design: it pairs a different vision encoder with the newer Qwen3 language backbone, injects multi-level visual features into multiple LLM layers via DeepStack, and replaces the axis-partitioned M-RoPE with an interleaved variant~\cite{qwen3vl}.
Visual Tracker Heads are identified with the same automatic two-stage procedure as in Section~\ref{sec:vth}, which requires only forward passes on 1,000 samples, and all other hyperparameters are kept unchanged.
Table~\ref{tab:qwen3_results} reports the results; deltas are measured relative to the Qwen3-VL-4B-Instruct base model.

GRIT improves OCR-Reasoning, GRIT-Bench Accuracy, and RF, whereas GRPO and DAPO do not improve over the base model in this setting.
These results indicate that GRIT's benefit is not tied to a specific architecture: the same automatic procedure, with unchanged hyperparameters, transfers across two architecturally distinct VLM generations.

\begin{table}[t]
    \centering
    \caption{Results on Qwen3-VL-4B-Instruct. Values in parentheses denote changes from the base model. Best results are in bold.}
    \label{tab:qwen3_results}
    \resizebox{\columnwidth}{!}{%
    \begin{tabular}{lccc}
        \toprule
        \multirow{2}{*}{\textbf{Model}} & \multirow{2}{*}{\textbf{OCR-Reasoning}} & \multicolumn{2}{c}{\textbf{GRIT-Bench}} \\
        \cmidrule(lr){3-4}
        & & \textbf{Acc.} & \textbf{RF} \\
        \midrule
        Qwen3-VL-4B-Instruct (base)
        & 32.27 & 71.1 & 80.3 \\

        GRPO-4B
        & 31.99 \scriptsize\textcolor{red}{($-0.28$)}
        & 69.4 \scriptsize\textcolor{red}{($-1.7$)}
        & 78.1 \scriptsize\textcolor{red}{($-2.2$)} \\

        DAPO-4B
        & 32.09 \scriptsize\textcolor{red}{($-0.18$)}
        & 70.9 \scriptsize\textcolor{red}{($-0.2$)}
        & 78.3 \scriptsize\textcolor{red}{($-2.0$)} \\

        \rowcolor{gray!20}
        \textbf{GRIT-4B (Ours)}
        & \textbf{36.01 \scriptsize\textcolor{green!50!black}{($+3.74$)}}
        & \textbf{77.5 \scriptsize\textcolor{green!50!black}{($+6.4$)}}
        & \textbf{83.9 \scriptsize\textcolor{green!50!black}{($+3.6$)}} \\
        \bottomrule
    \end{tabular}%
    }
\end{table}

\end{document}